\documentclass[journal]{IEEEtran}
\usepackage{cite}

\ifCLASSINFOpdf
\else
\fi
\usepackage[cmex10]{amsmath}
\usepackage{array}
\usepackage{url}

\usepackage{times}
\usepackage{epsfig}
\usepackage{graphicx}
\usepackage{amsthm,amsmath,amssymb,latexsym,epsfig}
\usepackage{multirow}
\usepackage{mathtools}
\usepackage{color}

\definecolor{col0}{RGB}{0,128,0}
\definecolor{col1}{RGB}{177,105,34}
\definecolor{col2}{RGB}{255,255,55}
\definecolor{col3}{RGB}{255,255,143}
\definecolor{col4}{RGB}{255,255,227}
\definecolor{col5}{RGB}{255,73,209}
\definecolor{col6}{RGB}{255,0,0}
\definecolor{col7}{RGB}{128,0,64}
\definecolor{col8}{RGB}{22,25,150}
\definecolor{col9}{RGB}{0,0,128}

\graphicspath{{Figures/}}

\begin{document}
%
\title{Deep Multi-task Learning \\for Railway Track Inspection}
%
%
%

\author{Xavier~Gibert,~\IEEEmembership{Student Member,~IEEE,}
        Vishal~M.~Patel,~\IEEEmembership{Member,~IEEE,}
        and~Rama~Chellappa,~\IEEEmembership{Fellow,~IEEE}
}

\maketitle

\begin{abstract}

Railroad tracks need to be periodically inspected and monitored to ensure safe transportation. Automated track inspection using computer vision and pattern recognition methods have recently shown the potential to improve safety by allowing for more frequent inspections while reducing human errors. Achieving full automation is still very challenging due to the number of different possible failure modes as well as the broad range of image variations that can potentially trigger false alarms. Also, the number of defective components is very small, so not many training examples are available for the machine to learn a robust anomaly detector. In this paper, we show that detection performance can be improved by combining multiple detectors within a multi-task learning framework. We show that this approach results in better accuracy in detecting defects on railway ties and fasteners.

\end{abstract}


%
\IEEEpeerreviewmaketitle

\section{Introduction}
\begin{figure}[t]
\begin{center}
  \includegraphics[trim=15mm 15mm 15mm 20mm, clip=true, width=3.5in]{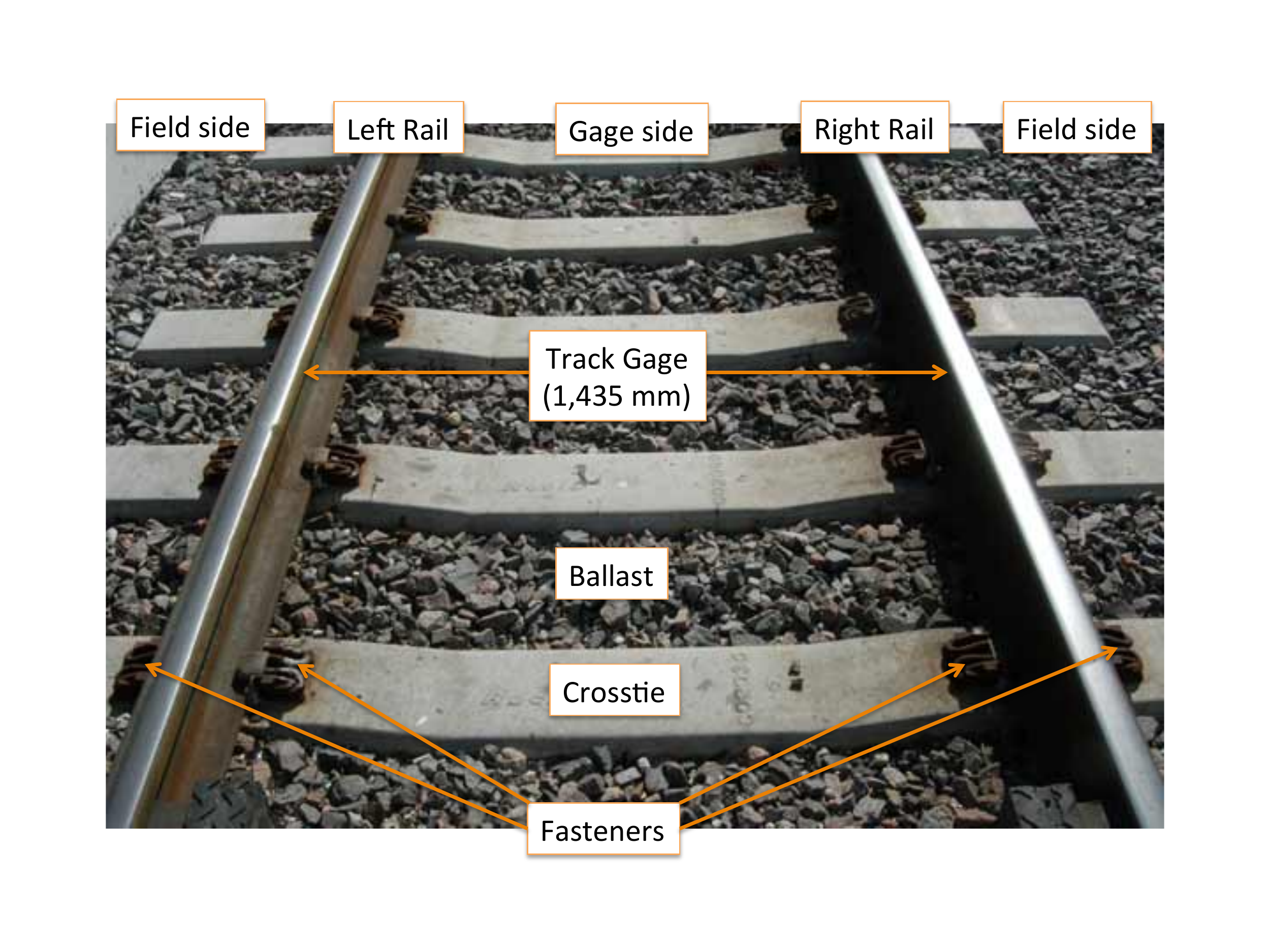}
\end{center}\vskip-20pt
\caption{ \label{fig:definitions}
Definition of basic track elements.}
\end{figure}

%
%
%
%
\IEEEPARstart{M}{onitoring} the condition of railway components is essential to ensure train safety, especially on High Speed Rail (HSR) corridors.  Amtrak's recent experience with concrete ties has shown that they have different kind of problems than wood ties \cite{S12}. The locations and names of the basic
track elements mentioned in this paper are shown in Figure~\ref{fig:definitions}.   Although concrete ties have life expectancies of up to 50 years, they may fail prematurely for a variety of reasons, such as the result of alkali-silicone reaction (ASR)  \cite{ST00} or delayed ettringite formation  \cite{ST04}. ASR is a chemical reaction between cement alkalis and non-crystalline (amorphous) silica. This forms alkali-silica gel at the aggregate surface. These reaction rims have a very strong affinity with water and have a tendency to swell. These compounds can produce internal pressures that are strong enough to create cracks, allowing moisture to penetrate, and thus accelerating the rate of deterioration. Delayed Ettringite Formation (DEF) is a type of internal sulfate attack that occurs in concrete that has been cured at excessively high temperatures. In addition to ASR and DEF, ties can also develop fatigue cracks due to normal traffic or by being impacted by flying debris or track maintenance machinery. Once small cracks develop, repeated cycles of freezing and thawing will eventually lead to bigger defects.

Fasteners maintain gage by keeping both rails firmly attached to the crossties. According to the Federal Railroad Administration (FRA) safety database\footnote{http://safetydata.fra.dot.gov}, in 2013, out of 651 derailments due to track problems, 27 of them were attributed to gage widening caused by defective spikes or rail fasteners, and another 2 to defective or missing spikes or rail fasteners. Also, in the United States, regulations enforced by the FRA\footnote{49 CFR 213 -- Track Safety Standards} prescribe visual inspection of high-speed rail tracks with a frequency of once or twice per week, depending on track speed. These manual inspections are currently performed by railroad personnel, either by walking on the tracks or by riding a hi-rail vehicle at very low speeds. However, such inspections are subjective and do not produce an auditable visual record. In addition, railroads usually perform automated track inspections with specialized track geometry measurement vehicles at intervals of 30 days or less between inspections. These automated inspections can directly detect gage widening conditions. However, it is preferable to find fastening problems before they develop into gage widening conditions.

Recent advances in CMOS imaging technology, have resulted in commercial-grade line-scan cameras that are capable of capturing images at resolutions of up to 4,096$\times$2 and line rates of up to 140 KHz. At the same time, high-intensity LED-based illuminators with life expectancies in the range of 50,000 hours are commercially available. This technology enables virtually maintenance-free operation over several months. Therefore, technology that enables autonomous visual track inspection from an unattended vehicle (such as a passenger train) may become a reality in the not-too-distant future. In previous works \cite{GP14fast,GP15icip} we showed that it is possible to automatically inspect the condition of ties and fasteners. In this paper, we extend these techniques and integrate them in a multi-task learning framework. This combined system achieves better performance than learning each task separately.

\subsection{Organization of the paper}
This paper is organized as follows. Related works on inspection of railway tracks using computer vision are discussed in section \ref{sec:prior_work}. The problem addressed in this paper is described in section \ref{sec:problem_desc}. Overall approach and system architecture is presented in section \ref{sec:approach}. Material classification, segmentation and tie assessment algorithm is described in section \ref{sec:material}. Fastener detection and assessment algorithm is described in section \ref{sec:fastener}. Experimental results are presented in section \ref{sec:experiments}, and section \ref{sec:conclusion} concludes the paper with a brief summary and discussion.


\section{Related Works}\label{sec:prior_work}

\subsection{Railway Track Inspection}

Since the pioneering work by Trosino \emph{et al.} \cite{AMK00,AMK02}, machine vision technology has been gradually adopted by the railway industry as a track inspection technology. Those first generation systems were capable of collecting images of the railway right of way and storing them for later review, but they did not facilitate any automated detection. As faster processing hardware became available, several vendors began to introduce vision systems with increasing automation capabilities.

In \cite{MD07,deruvo2009gpu}, Marino \emph{et al.} describe their VISyR system, which detects hexagonal-headed bolts using two 3-layer neural networks (NN) running in parallel. Both networks take the 2-level discrete wavelet transform (DWT) of a 24$\times$100 pixel sliding window (their images use non-square pixels) as an input to generate a binary output indicating the presence of a fastener. The difference is that the first NN uses Daubechies wavelets, while the second one uses Haar wavelets. This wavelet decomposition is equivalent to performing edge detection at different scales with two different filters. Both neural networks are trained with same examples. The final decision is made using the maximum output of each neural network. 

In \cite{GB07,BN08}, Gibert \emph{et al.} describe their VisiRail system for joint bar inspection. The system is capable of collecting images on each rail side, and finding cracks on joint bars using edge detection and a Support Vector Machine (SVM) classifier that analyzes features extracted from these edges. In \cite{PB09}, Babenko describes a fastener detection method based on a convolutional filter bank that is applied directly to intensity images. Each type of fastener has a single filter associated with it, whose coefficients are calculated using an illumination-normalized version of the Optimal Tradeoff Maximum Average Correlation Height (OT-MACH) filter \cite{M94}. This approach allowed accurate fastener detection and localization and achieved over 90\% fastener detection rate on a dataset of 2,436 images. However, the detector was not tested on longer sections of track. In \cite{RHA13}, Resendiz \emph{et al.}  use texture classification via a bank of Gabor filters followed by an SVM to determine the location of rail components such as crossties and turnouts. They also use the MUSIC algorithm to find spectral signatures to determine expected component locations. In \cite{LTHOP14}, Li \emph{et al.} describe a system for detecting tie plates and spikes. Their method, which is described in more detail in \cite{LOHP12enh}, uses an AdaBoost-based object detector \cite{VJ01} with a model selection mechanism that assigns the object class that produces the highest number of detections within a window of 50 frames. Table \ref{table:taxonomy_rail} summarizes several systems reported in the literature.
\begin{table*}[htp!]
\caption{Evolution of automated visual railway component inspection methods.}
\small
\label{table:taxonomy_rail}
\begin{center}
\begin{tabular}{c | c | c | c | c | c }
  \hline
  Authors & Year & Components & Defects & Features & Decision methods \\
  \hline
  \hline
  Stella \emph{et al.} \cite{stella2002,marino2007real,deruvo2009gpu} & 2002--09 & Fasteners & Missing & DWT & 3-layer NN \\
  Singh \emph{et al.} \cite{SS06} & 2006 & Fasteners & Missing & Edge density & Threshold \\
  Hsieh \emph{et al.} \cite{HC07} & 2007 & Fasteners & Broken & DWT & Threshold \\
  Gibert \emph{et al.} \cite{GB07,BN08} & 2007--08 & Joint Bars & Cracks & Edges & SVM \\
  Babenko \cite{PB09} & 2008 & Fasteners & Missing/Defective & Intensity & OT-MACH corr. \\
  Xia \emph{et al.} \cite{XX10} & 2010 & Fasteners & Broken & Haar & Adaboost \\
  Yang \emph{et al.} \cite{YQ11} & 2011 & Fasteners & Missing & Direction Field & Correlation \\
  Resendiz \emph{et al.} \cite{RHA13} & 2013 & Ties/Turnouts & -- & Gabor & SVM\\
  Li \emph{et al.} \cite{LTHOP14} & 2014 & Tie plates & Missing spikes & Lines/Haar & Adaboost \\
  Feng \emph{et al.} \cite{FG14} & 2014 & Fasteners & Missing/Defective & Haar & PGM \\
  Gibert \emph{et al.} \cite{GP14} & 2014 & Concrete ties & Cracks & DST & Iterative shrinkage \\
  Khan \emph{et al.} \cite{KI14} & 2014 & Fasteners & Defective & Harris-Stephen, Shi-Tomasi & Matching errors \\
  Gibert \emph{et al.} \cite{GP14fast} & 2015 & Fasteners & Missing/Defective & HOG & SVM \\
  Gibert \emph{et al.} \cite{GP15icip} & 2015 & Concrete ties & Tie Condition & Intensity & Deep CNN \\
  \hline
\end{tabular}
\end{center}
\end{table*}

\subsection{Convolutional Neural Networks}

The idea of enforcing translation invariance in neural networks via weight sharing goes back to Fukoshima's Neocognitron\cite{F80}. Based on this idea, LeCun \emph{et al.} developed the concept into Deep Convolutional Neural Networks (DCNN) and used it for digit recognition\cite{LB89}, and later for more general optical character recognition (OCR)\cite{LB98}. During the last few years, DCNNs have become ubiquitous in achieving state-of-the-art results in image classification\cite{KSH13,SL14} and object detection \cite{GD14}. This resurgence of DCNNs has been facilitated by the availability of efficient GPU implementations and open source libraries such as Caffe\cite{caffe} and Torch7\cite{torch7}. More recently, DCNNs have been used for semantic image segmentation. For example, the work of \cite{LSD14} shows how a DCNN can be converted to a Fully Convolutional Network (FCN) by replacing fully-connected layers with convolutional ones.

\subsection{Multi-task Learning}

Multi-task learning (MTL) is an inductive transfer learning technique in which two or more learning machines are trained cooperatively\cite{RC97}. It is a generalization of multi-label learning in which each training sample has only been labeled for one of the tasks. In MTL settings there is a mechanism in which knowledge learned for one task is transferred to the other tasks\cite{PMK91}. The idea is that each task can benefit by reusing knowledge that has been learned while training for the other tasks. Backpropagation has been recognized as an effective method for learning distributed representations \cite{GH86}. For instance, in multitask neural networks, we jointly minimize one global loss function
\begin{align}
\Phi = \sum_{t=1}^T \lambda_t \sum_{i=1}^{N_t} E_t \left( f(x_{ti}),y_{ti} \right),
\end{align}
where $T$ is the number of tasks, $N_t$ is the number of training samples for task $t$, $y_{ti}$ is the ground truth label for training sample $x_{ti}$, $f$ is the the multi-output function computed by the network, and $E_t$ is the loss function for task $t$. This contrasts with the Single Task Learning (STL) setting, in which we minimize $T$ independent loss functions
\begin{align}
\Phi_t = \sum_{i=1}^{N_t} E_t \left( f_t(x_{ti}),y_{ti} \right), \quad t \in \{ 1 \dots T\}.
\end{align}

In MTL, the weighting factor $\lambda_t$ is necessary to compensate for imbalances in the complexity of the different tasks and the amount of training data available. When using back-propagation, it is necessary to adjust $\lambda_t$'s to ensure that all tasks are learning at optimal rates.

\subsection{One-shot Learning}

To achieve good generalization performance, traditional machine learning methods require a minimum number of training examples from each class. This is necessary for the machine to learn a model that can handle variations in image appearance that result from changes in illumination, scale, rotation, background clutter, and so on. However, the occurrence of each type of anomaly is very infrequent, so in anomaly detection settings it is only possible to find one or a few number of examples from which to learn from. If we try to learn a complete model for a new class using such a limited number of examples, this model would overfit and would not be able to generalize to new data. However, if we reuse knowledge that has been learned while learning other related classes, we can learn better models. This is known as one-shot learnig\cite{FFP06}. We pose this one-shot learning problem as a special case of multi-task learning, in which one task consists of learning the abundant classes, while the other task learns the uncommon classes. Both tasks share a common low-level representation because all fasteners are built with common materials. In this paper, we train an auxiliary network on a 5-class fastener classification using more than 300K fasteners for the sole purpose of learning a good representation that regularizes the broken fastener detector.


\section{Problem Description}\label{sec:problem_desc}

The application described in this paper consists of inspecting railroad tracks for defects on crossties and rail fasteners using single-view line-scan cameras. The crossties may of different materials (e.g. wood, concrete, plastic, or metal), and the fasteners could be of different types (e.g. elastic clips, bolts, or spikes). We have posed this problem as two detection problems: object detection (good, broken, or missing fastener), and semantic segmentation (chips and crumbling concrete ties and other material classes).

\subsection{Dataset}

The dataset used to demonstrate this approach comprises 85 miles of track in which the bounding boxes of 203,287 ties have been provided. This data is very challenging to work with. The images were collected from a moving vehicle and although there was artificial illumination, there are significant variations in illumination due to sun position and shadows. To reduce friction between rails and wheels and prolong their usable lives, railroads may lubricate them using special equipment mounted along the tracks. At locations near these lubricators, tracks get dirty and the accumulation of greasy deposits significantly change the appearance of the images. There are also some spots in which the tracks are covered by mud being pumped through the ballast during heavy rainfall. Moreover, there are also places in which the ballast is unevently distributed and pieces of ballast rock cover the ties and fasteners being inspected. Also, leaves, weeds, branches, trash and other debris may occlude the track components being inspected.

\subsection{Data Annotation}

\begin{figure}[t!]
\begin{center}
\begin{tabular}{c}
  \includegraphics[width=3.5in]{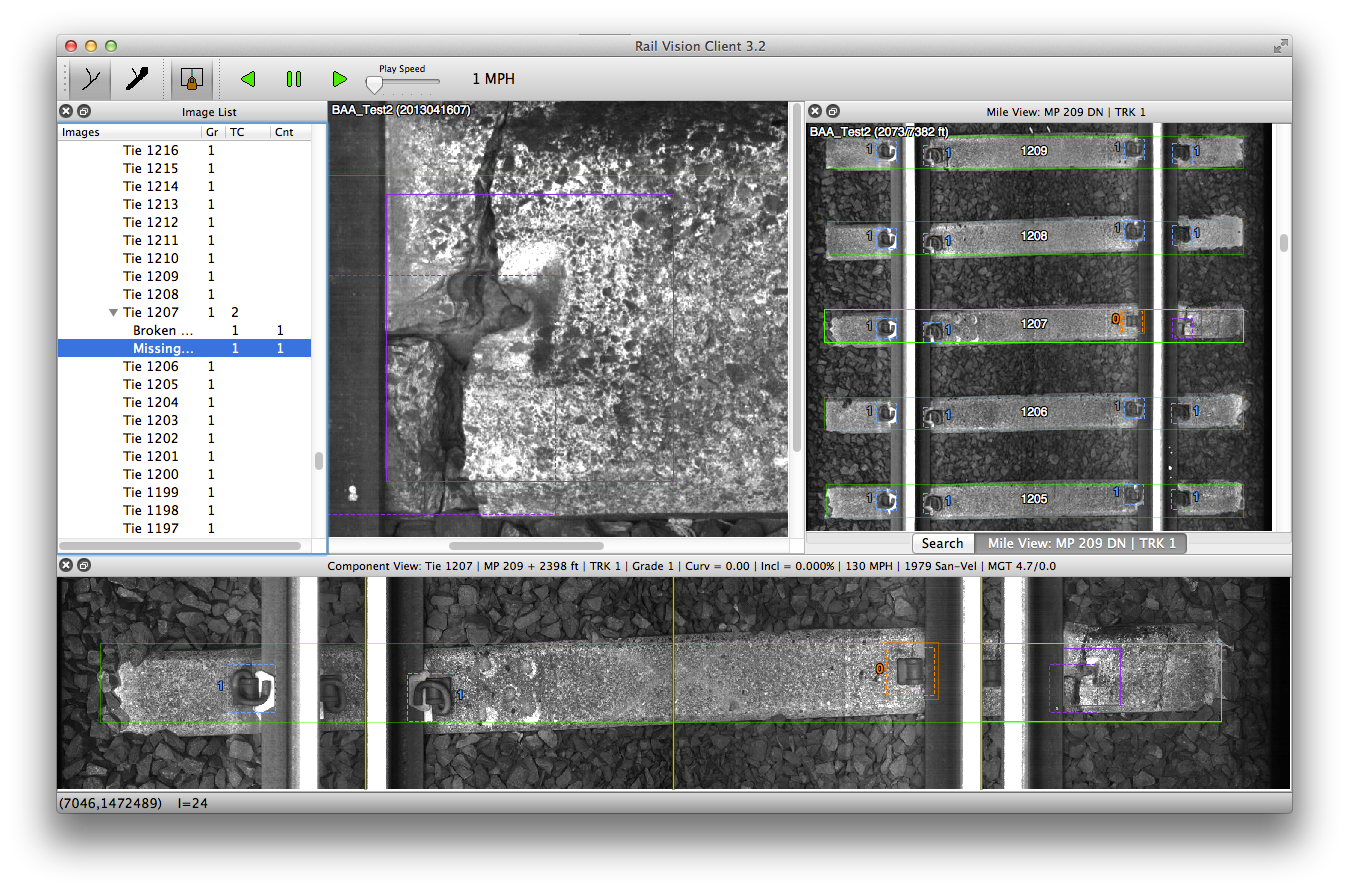} \\
  \includegraphics[width=3.5in]{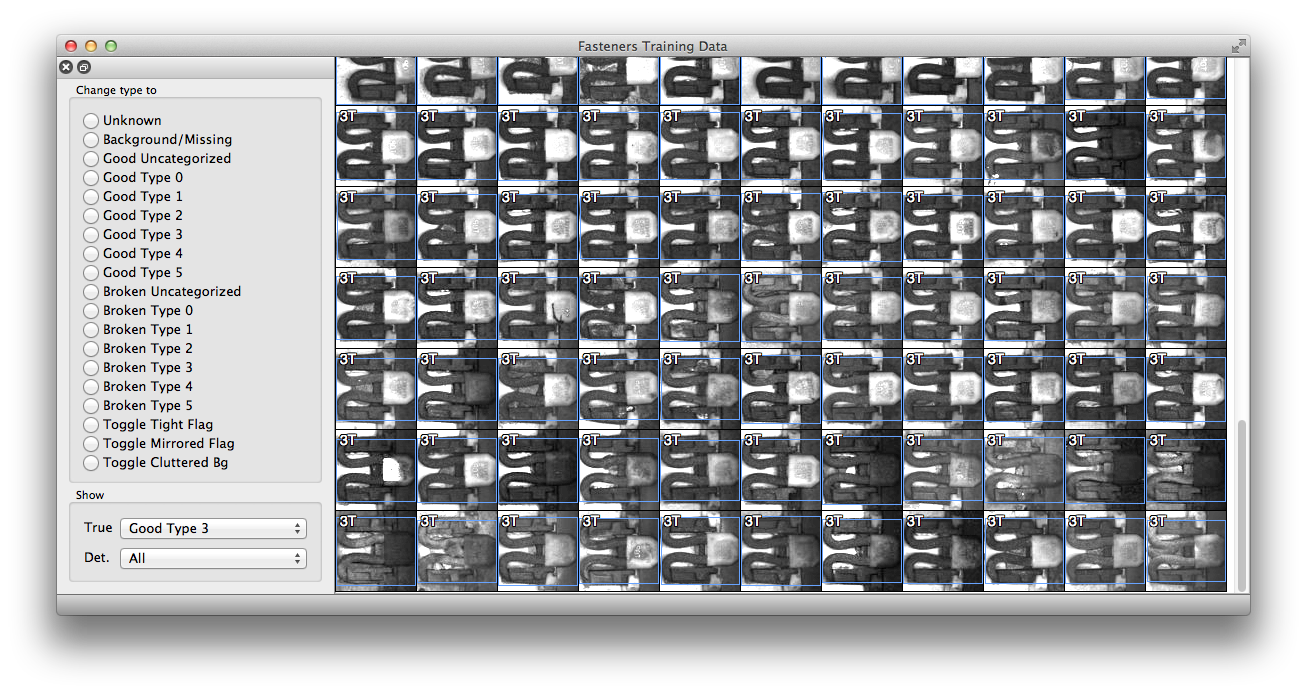} \\
\end{tabular}
\end{center}
\caption
{ \label{fig:gui}
GUI tool used to generate the training set and to review the detection results.
}
\end{figure}

Due to the large size of this dataset, we have implemented a customized software tool that allows the user to efficiently visualize and annotate the data (see Figure \ref{fig:gui} for a screenshot). This tool has been implemented in C++ using the Qt framework and communicates with the data repository through the secure HTTPS protocol, so it can be used from any computer with an Internet connection without having to set up tunnel or VPN connections. The tool allows assigning a material category to each tie as well as its bounding box. The tool also allows defining polygons enclosing regions containing crumbling, chips or ballast. The tool also allows the user to change the threshold of the defect detector and select a subset of the data for display and review. It also has the capability of exporting lists of detected defects as well as summaries of fastener inventories by mile.


\section{Approach}\label{sec:approach}

\subsection{Overall Architecture}\label{sec:arch}

\begin{figure*}[t]
\begin{center}
  \includegraphics[trim=0mm 115mm 0mm 0mm, clip=false, width=7in]{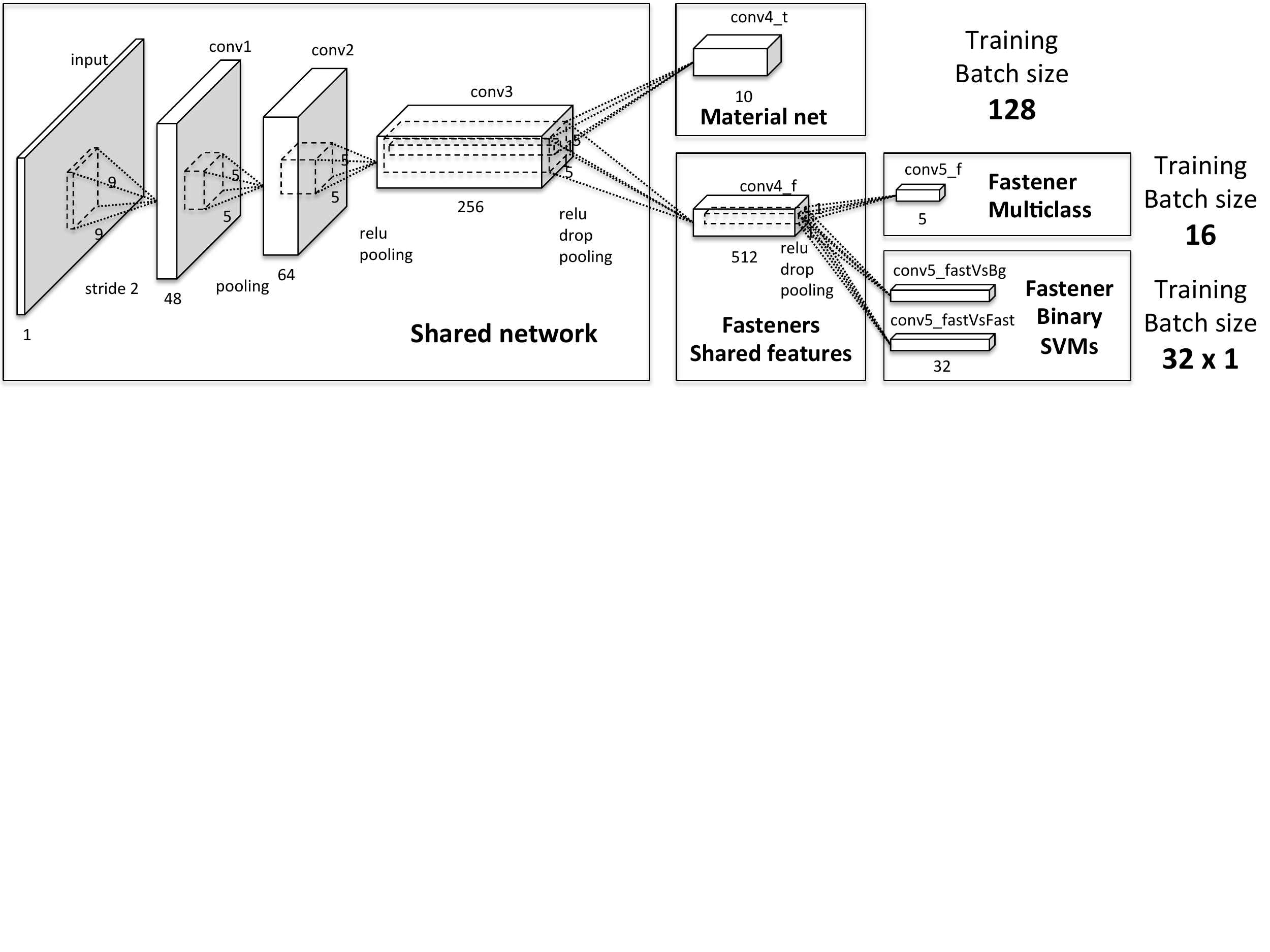}
\end{center}
\caption
{ \label{fig:architecture}
Network architecture. }
\end{figure*}

Our design is a Fully Convolutional Network\cite{LSD14} based on the architecture that we introduced in \cite{GP15icip}. That network was trained with 10 classes of materials and produces feature maps with 10 different channels. In this paper, we extend that architecture by adding two additional branches to the network. The first one is a coarse-level fastener classifier trained on a large number of examples. The second branch produces 32 binary outputs. These outputs correspond to the same binary SVMs that we used in our previous version of the detector\cite{GP14fast} described in more detail in section \ref{sec:fastener}.

The implementation is based on the BVLC Caffe framework\cite{caffe}. For the material classification task, we have a total of 4 convolutional layers between the input and the output layer, while for fastener detection tasks we have 5 convolutional layers. The first three layers are shared among all the tasks. The fasteners task is, in turn, divided in two subtasks: coarse-level and fine-grained classification (see section \ref{sec:fastener} for more details). The network uses rectified linear units (ReLU) as non-linear activation functions, and overlapping max pooling units of size $3 \times 3$. All max pooling units have a stride of 2, except the one on top of that has a stride of 1. We use dropout\cite{dropout} regularization on layer 3 (with a ratio of 0.1) and layer 4 on the fasteners branch (with a ratio of 0.2). The network also uses weight decay regularization. On the fasteners branch, we increase the weight decay factors on layers 4 and 5 by $10\times$ and $100\times$ respectively to reduce overfitting.

We first apply global gain normalization on the raw image to reduce the intensity variation across the image. This gain is calculated by smoothing the signal envelope estimated using a median filter.  We estimate the signal envelope by low-pass filtering the image with a Gaussian kernel. Although DCNNs are robust to illumination changes, normalizing the image to make the signal dynamic range more uniform improves accuracy and convergence speed. We also subtract the mean intensity value, which is calculated on the whole training set. The network architecture is illustrated in Figure \ref{fig:architecture}.

\subsection{Training Procedure}

To generate our training set, we initially selected $\sim$30 good quality (with no occlusion and clean edges) samples from each object category at random from the whole repository and annotated the bounding box location and object class for each of them. Our training software also automatically picks, using a randomly generated offset, a background patch adjacent to each of the selected samples. Once we had enough samples from each class, we trained binary classifiers for  each of the classes against the background and tested on the whole dataset. Then, we randomly selected misclassified samples and added those that had good or acceptable quality (no occlusion) to the training set. To maintain the balance of the training set, we also added, for each difficult sample, 2 or 3 neighboring samples. Since there are special types of fasteners that do not occur very frequently (such as the c-clips or j-clips used around joint bars), in order to keep the number of samples of each type in the training set as balanced as possible, we added as many of these infrequent types as we could find.

Careful annotation of the dataset resulted in the training set of 2819 fully-annotated fasteners.  
Moreover, some of the classes had very few examples. For instance, there are only 28 broken fast-clips,  and just 38 j-clips in the dataset. 
If we just had used this limited data, we would not have been able to learn a good representation. Fortunately, both of these two uncommon classes of fasteners share parts with the other ones. Therefore, if we can make layer \emph{conv4\_f} learn a good model for fastener parts, layer \emph{conv5\_f} would be able to learn how to distinguish between fasteners by combining such parts, even if the number of training examples is limited.

Therefore, we created an auxiliary fastener data set. Since the only purpose of this dataset is to help learn parts, we just used the bounding boxes and labels automatically generated by our previous detector\cite{GP14fast}, whose error rate is just 0.37\%. We sampled 62,500 fasteners from each of 5 coarse classes. The first class contains missing and broken fasteners, the next 3 classes contain fasteners corresponding to each of the classes containing the most samples (PR-clips, e-clips, and fast-clips), and the last class contains everything else.

We train the network using stochastic gradient descent on mini-batches of 128 image patches of size $75 \times 75$ plus 48 fastener images of $182 \times 182$. The fastener images include 16 from the auxiliary fastener dataset and 1 from each of the binary SVM tasks. We do data augmentation on material classification by randomly mirroring vertically and/or horizontally the training samples. The patches are cropped randomly among all regions that contain the texture of interest. To increase robustness against adverse environment conditions, such as rain, grease or mud, we identified images containing such difficult cases and automatically resampled the data so that at least 50\% of the data is sampled from such difficult images. We do data augmentation on fasteners by randomly mirroring vertically the symmetric classes and randomly cropping the fastener offset uniformly distributed within a +/-9 pixel range in both directions.


\section{Material Identification and Segmentation}\label{sec:material}

\subsection{Architecture}

\begin{figure}[t]
\begin{center}
\begin{tabular}{c c c}
  \fcolorbox{col0}{col0}{\includegraphics[width=14mm]{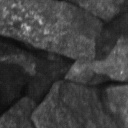}} &
  \fcolorbox{col1}{col1}{\includegraphics[width=14mm]{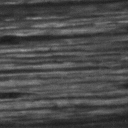}} &
  \fcolorbox{col2}{col2}{\includegraphics[width=14mm]{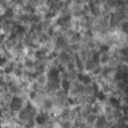}} \\
  (a) & (b) & (c) \\
\end{tabular}
 
\vspace{3mm}
\begin{tabular}{c c c c}
  \fcolorbox{col3}{col3}{\includegraphics[width=14mm]{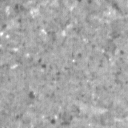}} &
  \fcolorbox{col4}{col4}{\includegraphics[width=14mm]{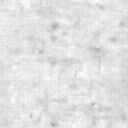}} &
  \fcolorbox{col5}{col5}{\includegraphics[width=14mm]{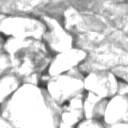}} &
  \fcolorbox{col6}{col6}{\includegraphics[width=14mm]{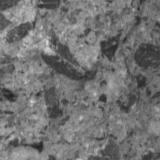}} \\
  (d) & (e) & (f) & (g) \\
\end{tabular}

\vspace{3mm}
\begin{tabular}{c c c}
  \fcolorbox{col7}{col7}{\includegraphics[width=14mm]{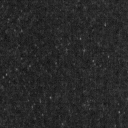}} &
  \fcolorbox{col8}{col8}{\includegraphics[width=14mm]{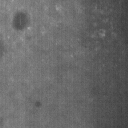}} &
  \fcolorbox{col9}{col9}{\includegraphics[width=14mm]{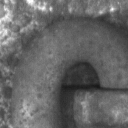}} \\
  (h) & (i) & (j)
\end{tabular}
\end{center}
\caption
{ \label{fig:texture_classes}
Material categories. (a) ballast (b) wood (c) rough concrete (d) medium concrete (e) smooth concrete (f) crumbling concrete (g) chipped concrete (h) lubricator (i) rail (j) fastener}
\end{figure}

The material classification task at layer \emph{conv4\_t} generates ten score maps at 1/16th. Each value $\Phi_i(x,y)$ in the score map corresponds to the likelihood that pixel location $(x,y)$ contains material of class $i$. The ten classes of materials are defined in Figure \ref{fig:texture_classes}.


\subsection{Score Calculation}\label{sec:score}

To detect whether an image contains a broken tie, we first calculate the scores at each site as
\begin{equation}
	S_b(x,y) = \max_{i \notin \mathcal{B}} \Phi_i(x,y) - \Phi_b(x,y)
\end{equation}
where $b \in \mathcal{B}$ is a defect class (crumbling or chip). Then we calculate the score for the whole image as
\begin{equation}
\label{eq:scoret}
	S_b = \frac{1}{\beta - \alpha} \int_\alpha^\beta \widehat{F}^{-1}(t) dt
\end{equation}
where $\widehat{F}^{-1}$ refers to the $t$ sample quantile calculated from all scores $S_b(x,y)$ in the image. The detector reports an alarm if $S > \tau$, where $\tau$ is the detection threshold. We used $\alpha=0.9$ and $\beta=1$.


\section{Fasteners Assessment}\label{sec:fastener}

\begin{figure}[htp!]
\begin{center}
  \includegraphics[width=3.25in]{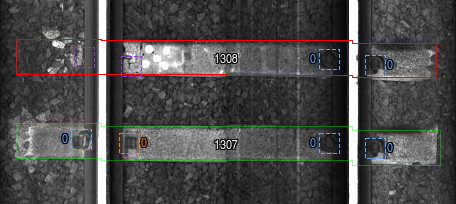}
\end{center}
\caption
{ \label{fig:cluster}
Example of defects that our algorithm can detect. Blue boxes indicate good fasteners, orange boxes indicate broken fasteners, and purple boxes indicate missing fasteners. White numbers indicate tie index from last mile post. Other numbers indicate type of fastener (for example, 0 is for e-clip fastener).}
\end{figure}

In this section, we describe the details of the fastener assessment task. Figure~\ref{fig:cluster} shows the types of defects that our algorithm can detect.

\subsection{Overview}

Due to surface variations that result from grease, rust and other elements in the outdoor environment, segmentation of railway components is a very difficult task. Therefore, we avoid it by using a detector based on a sliding window that we run over the inspectable area of the tie. The features learned at layer \emph{conv4\_f} are computed from the shared features at \emph{conv3}. The reason for sharing the features with the material classification task is that there is overlap between both tasks. For instance, the material classification task needs to learn to distinguish between fasteners and the other materials, regardless of the type of fastener. Also, the fastener detection class needs to discriminate between fasteners and background, regardless of the type of background. In our previous work, we used the Histogram of Oriented Gradients (HOG) \cite{DT05} descriptor for detecting fasteners. Although the results that we obtained using HOG features were better than previously proposed methods, this approach still has its limitations. For instance, the dimensionality of the feature vector is very large (12,996), consuming a lot of memory and computational resources, and the linear classifier cannot handle occlusions very well. Therefore, in this paper we attempt to learn the features by training the network end to end.

\begin{figure*}[htp!]
\begin{center}
  \includegraphics[trim=0 4.25cm 0 0, clip=true, width=5in]{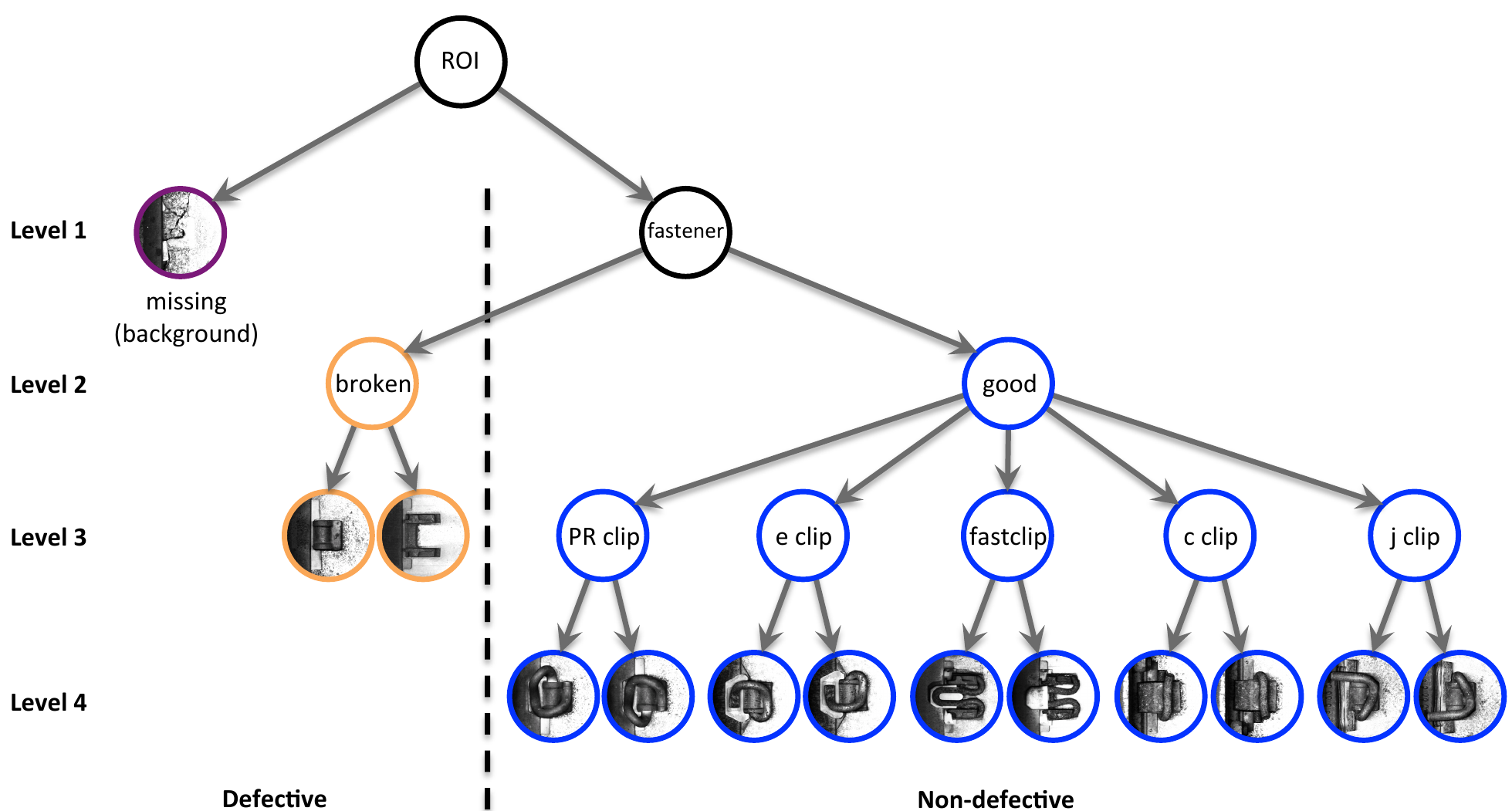}
\end{center}
\caption
{ \label{fig:fastener_classes}
Object categories used for detection and classification (from coarsest to finest levels).}
\end{figure*}

\subsection{Classification}
\label{sec:classification}

Our goal is to simultaneously detect, within each predefined Region of Interest (ROI), the most likely fastener location and then classify such detections into one of three basic conditions: background (or missing fastener), broken fastener, and good fastener. Then, for good and broken fastener conditions, we want to assign class labels for each fastener type (PR clip, e-clip, fastclip, c-clip, and j-clip). Figure \ref{fig:fastener_classes} shows the complete categorization that we use, from coarsest to finest. At the coarsest level, we want to classify fastener vs. unstructured background clutter. The background class also includes images of ties where fasteners are completely missing. We have done this for these reasons: 1) it is very difficult to train a detector to find the small hole left on the tie after the whole fastener has been ripped off, 2) we do not have enough training examples of missing fasteners, and 3) most missing fasteners are on crumbled ties for which the hole is no longer visible. Once we detect the most likely fastener location, we want to classify the detected fastener between broken vs. good, and then classify it into the most likely fastener type. Although this top-down reasoning works for a human inspector, it does not work accurately in a computer vision system because both the background class and the fastener class have too much intra-class variations.  As a result, we have resorted to a bottom-up approach.

To achieve the best possible generalization at test time, we have based our detector on the maximum margin principle of the SVM. The SVM separating hyperplane is obtained by minimizing the regularized hinge loss function,
\begin{align}
E = \sum_i \max\left( 0, 1-y_i (w \cdot x_i+b) \right) + \frac{\lambda}{2} \| w \|,
\end{align}
where $x_i \in \mathbb{R}^{512}$ are the outputs of layer \emph{conv4\_f} and $y_i \in \{-1,+1\}$ their corresponding ground truth labels (whose meaning will be explain later). The gradients with respect to the parameters $w$ and $b$ are
\begin{align}
\frac{\partial E}{\partial w} &= - \sum_i y_i x_i \delta[y_i (w \cdot x_i + b) < 1] + \lambda w \\
\frac{\partial E}{\partial b} &= - \sum_i y_i \delta[y_i (w \cdot x_i + b) < 1],
\end{align}
where $\delta\{\text{condition}\}$ is 1 if condition is true and -1 otherwise. The gradient of the hinge loss function with respect to the data (which is back-propagated down to the lower layers) is
\begin{align}
\frac{\partial E}{\partial x_i} = -y_i w \delta[y_i (w \cdot x_i + b) < 1].
\end{align}
Once the parameters converge, these gradients become highly sparse and only the difficult training samples contribute to to updating the parameters on layer \emph{conv4\_f} and all the layers below.

\begin{figure}[htp!]
\begin{center}
\begin{tabular}{c c}
  \includegraphics[trim=0 6cm 11cm 0cm, width=1.35in]{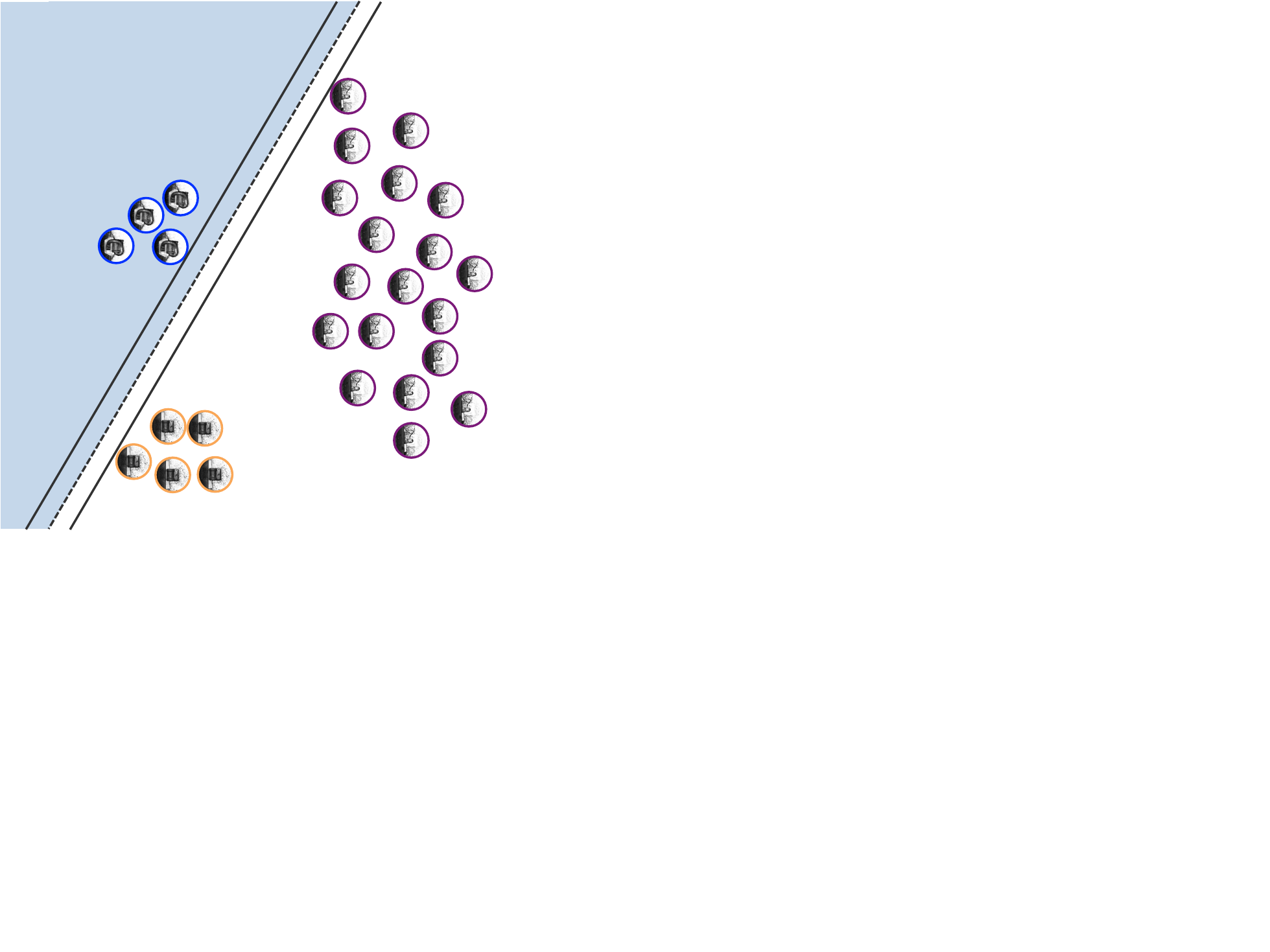} &
  \includegraphics[trim=0 6cm 11cm 0cm, width=1.35in]{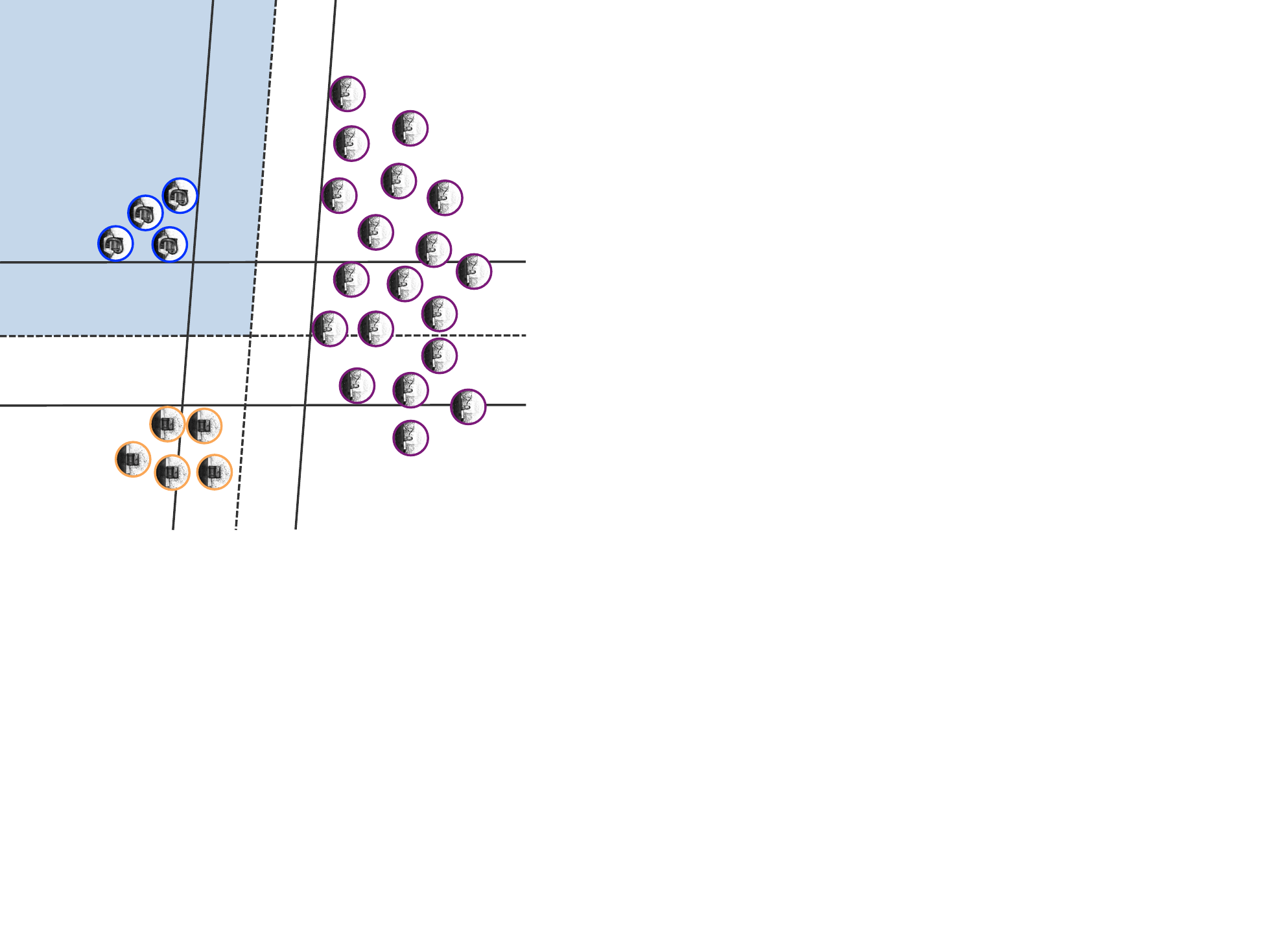} \\
  (a) & (b) \\
\end{tabular}
\end{center}
\caption
{ \label{fig:margin}
Justification for using two classifiers for each object category. Shaded decision region corresponds fastener in good condition, while white region corresponds to defective fastener. Blue circles are good fasteners, orange circles are broken fasteners, and purple circles are background/missing fasteners.  (a) Classification region of good fastener vs. rest (b) Classification region of intersection of good fastener vs. background and good fastener vs. rest-minus-background. The margin is much wider than using a single classifier.
}
\end{figure}

Instead of training a multi-class SVM, we use the one-vs-rest strategy, but instead of treating the background class as just another object class, we treat it as a special case and use a pair of SVMs per object class. For instance, if we had used a single learning machine, we would be forcing the classifier to perform two different unrelated tasks: a) reject that the image patch does not contain random texture and b) reject that the object does not belong to the given category. Therefore, given a set of object classes $\mathcal{C}$, we train two detectors for each object category. The first one, with weights $b_c$, classifies each object class $c \in \mathcal{C}$ vs. the background/missing class $m \not\in \mathcal{C}$, and the second one, with weights $f_c$ classifies object class $c$ vs. other object classes $\mathcal{C} \backslash c$. As illustrated in Figure \ref{fig:margin}, asking our linear classifier to perform both tasks at the same time would result in a narrower margin than training separate classifiers for each individual task. Moreover, to avoid rejecting cases where all $f_c$ classifiers produce negative responses, but one or more $b_c$ classifiers produce strong positive responses that would otherwise indicate the presence of a fastener, we use the negative output of $f_c$ as a soft penalty. Then the likelihood that sample $x$ belongs to class $c$ becomes
\begin{align}
L_c(x) = b_c \cdot x + \min(0,f_c \cdot x),
\end{align}
where $x=HOG(I)$ is the feature vector extracted from a given image patch $I$.  The likelihood that our search region contains at least one object of class $c$ is the score of the union, so
\begin{align}
\label{Lc}
L_c = \max\limits_{x\in\mathcal{X}} L_c(x),
\end{align}
where $\mathcal{X}$ is the set of all feature vectors extracted within the search region, and our classification rule becomes
\begin{align}
\hat c =
\begin{cases}
	\arg\max\limits_{c \in \mathcal{C}} L_c & \quad \max\limits_{c \in \mathcal{C}} L_c > 0 \\
	m & \quad \text{otherwise}.
\end{cases}
\end{align}

\subsection{Score Calculation}\label{sec:scoref}

For the practical applicability of our detector, it needs to output a scalar value that can be compared to a user-selectable threshold $\tau$. Since there are several ways for a fastener to be defective (either missing or broken), we need to generate a single score that informs the user how confident the system is that the image contains a fastener in good condition. This score is generated by combining the output of the binary classifiers introduced in the previous section.

We denote the subset of classes corresponding to good fasteners as $\mathcal{G}$ and that of broken fasteners as $\mathcal{B}$. These two subsets are mutually exclusive, so $\mathcal{C} = \mathcal{G} \cup \mathcal{B}$ and $\mathcal{G} \cap \mathcal{B} = \emptyset$. To build the score function, we first compute the score for rejecting the missing fastener hypothesis (i.e, the likelihood that there is at least one sample $x \in \mathcal{X}$ such that  $x \notin m$) as
\begin{align}
S_m = \max\limits_{c\in\mathcal{G}} L_c,
\end{align}
where $L_c$ is the likelihood of class $c$ defined in Eq. \refeq{Lc}. Similarly, we compute the score for rejecting the broken fastener hypothesis (i.e, the likelihood that for each sample $x \in \mathcal{X}$, $x \notin \mathcal{B}$  ) as
\begin{align}
S_b = -\max\limits_{c\in\mathcal{B}} \max\limits_{x\in\mathcal{X}} f_c \cdot x,
\end{align}
The reason why the $S_b$ does not depend on a $c$-vs-background classifier $b_c$ is because mistakes between missing and broken fastener classes do not need to be penalized. Therefore, $S_b$ need only produce low scores when $x$ matches at least one of the models in $\mathcal{B}$. The negative sign in $S_b$ results from the convention that a fastener in good condition should have a large positive score. The final score becomes the intersection of these two scores
\begin{align}
\label{eq:scoref}
S = \min(S_m, S_b).
\end{align}
The final decision is done by reporting the fastener as good if $S > \tau$, and defective otherwise.

\subsection{Training Procedure}
\label{sec:training}

The advantage of using a maximum-margin classifier is that once we have enough support vectors for a particular class, it is not necessary to add more inliers to improve classification performance. Therefore, we can potentially achieve relatively good performance with only having to annotate a very small fraction of the data.

\subsection{Alignment Procedure}

For learning the most effective object detection models, the importance of properly aligning the training samples cannot be emphasized enough. Misalignment between different training samples would cause unnecessary intra-class variation that would degrade detection performance. Therefore, all the training bounding boxes were manually annotated, as tightly as possible to the object contour by the same person to avoid inducing any annotation bias. For training the fastener vs. background detectors, our software cropped the training samples using a detection window centered around these boxes. For training the fastener vs. rest detectors, our software centered the positive samples using the user annotation, but the negative samples were re-centered to the position where the fastener vs. background detector generated the highest response. This was done to force the learning machine to learn to differentiate object categories by taking into account parts that are not common across object categories.


\section{Experimental Results}\label{sec:experiments}

\begin{figure}[t!]
\begin{center}
  \includegraphics[trim=0cm 0cm 0cm 0cm, height=3.1in]{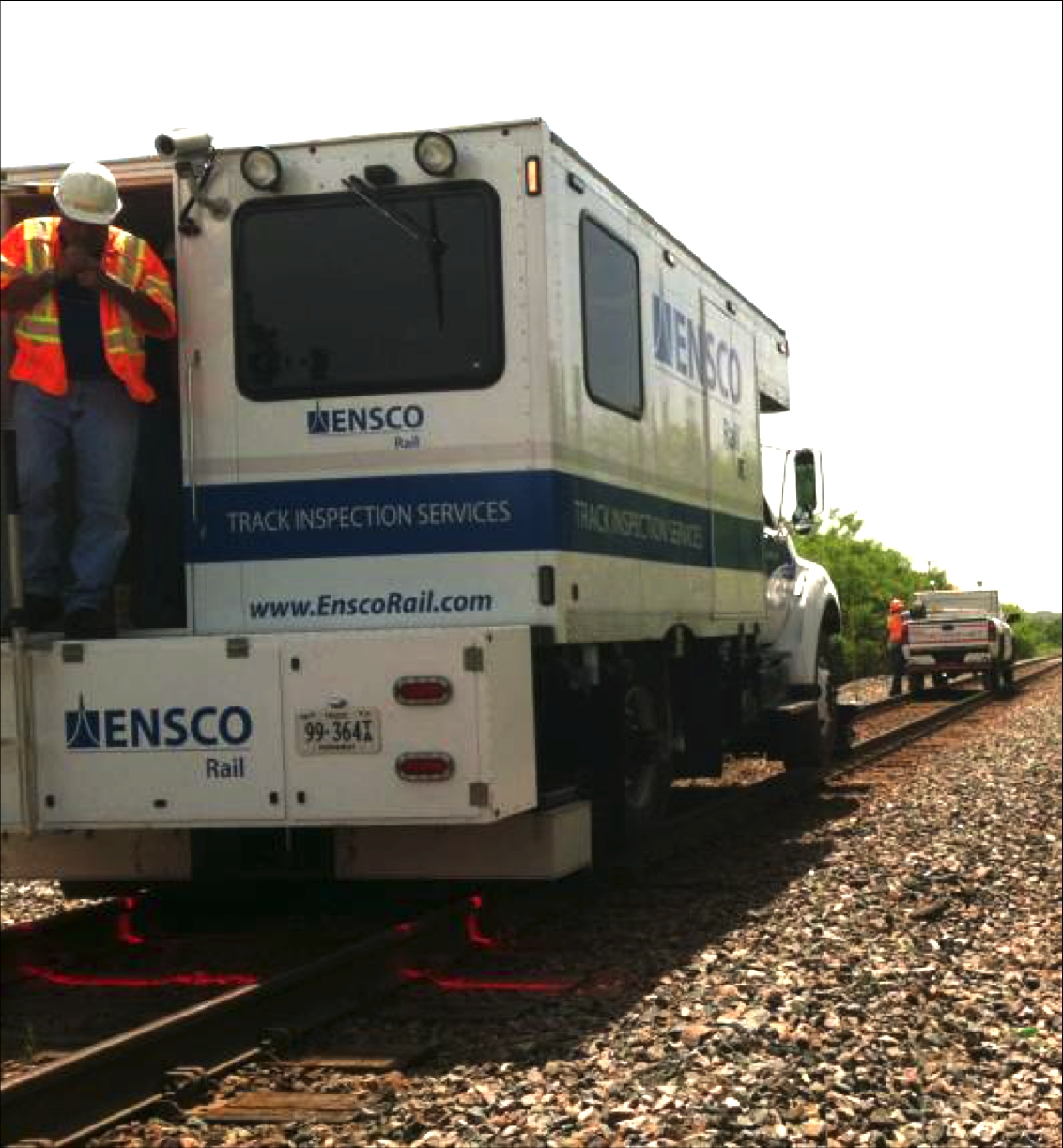}
\end{center}
\caption
{ \label{fig:ctiv}
CTIV platform used to collect the images.
}
\end{figure}
To evaluate the accuracy of our fastener detector, we have tested it on 85 miles of continuous trackbed images. These images were collected on the US Northeast Corridor (NEC) by ENSCO Rail's Comprehensive Track Inspection Vehicle (CTIV) (see Figure \ref{fig:ctiv}). The CTIV is a hi-rail vehicle (a road vehicle that can also travel on railway tracks) equipped with several track inspection technologies, including a Track Component Imaging System (TCIS). The TCIS collects images of the trackbed using 4 Basler sprint (spL2048-70km) linescan cameras and a custom line scan lighting solution\cite{B11}.

The sprint cameras are based on CMOS technology and use a CameraLink interface to stream the data to a rack-mounted computer. Each camera contains a sensor with 2 rows of 2,048 pixels that can sample at line rates of up to 70KHz. The cameras can be set to run in dual-line mode (high-resolution) or in ``binned'' more, where the values of each pair of pixels are averaged inside the sensor. During this survey, the cameras were set up in binned mode so, each camera generated a combined row of 2,048 pixels at a line rate of 1 line/0.43mm. The sampling rate was controlled by the signal generated from a BEI distance encoder mounted on one of the wheels. The camera positions and optics were selected to cover the whole track with minimal perspective distortion and their fields of view had some overlap to facilitate stitching.

The collected images were automatically stitched together and saved into several files, each containing a 1-mile image. These files were preprocessed by ENSCO Rail using their proprietary tie detection software to extract the boundary of all the ties in each image. We verified that the tie boundaries were accurate after visually correcting invalid tie detections. We downsampled the images by a factor of 2, for a pixel size of 0.86 mm. To assess the detection performance under different operating conditions, we flagged special track sections where the fastener visible area was less than 50\% due to a variety of occluding conditions, such as protecting covers for track-mounted equipment or ballast accumulated on the top of the tie. We also flagged turnouts so we could report separate ROC curves for both including and excluding them. All the ties in this dataset are made of reinforced concrete, were manufactured by either San-Vel or Rocla, and were installed between 1978 and 2010.

For a fair comparison between the approach proposed in this paper and previously published results, we trained the algorithm with the same dataset and annotations that we used in our previous works described in \cite{GP15icip} and \cite{GP14fast}. We used the output of our previous fastener detection algorithm\cite{GP14fast} to extract new fastener examples for semisupervised learning.


\begin{figure*}[th]
\begin{center}
\begin{tabular}{c c}
  \includegraphics[trim=0mm 0mm 38mm 0mm, clip=true, width=3.7in]{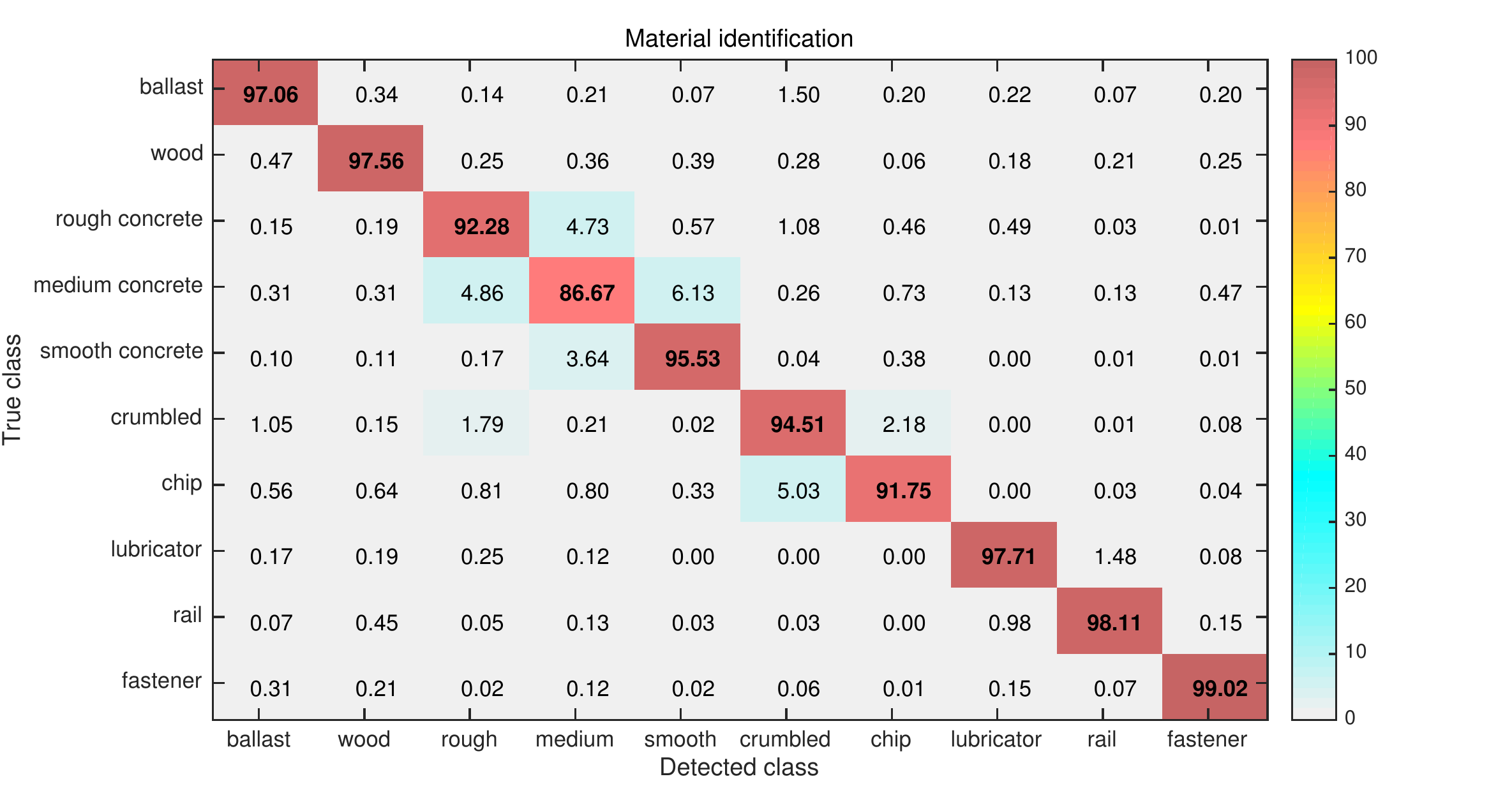} &
  \includegraphics[trim=33mm 0mm 3mm 0mm, clip=true, width=3.7in]{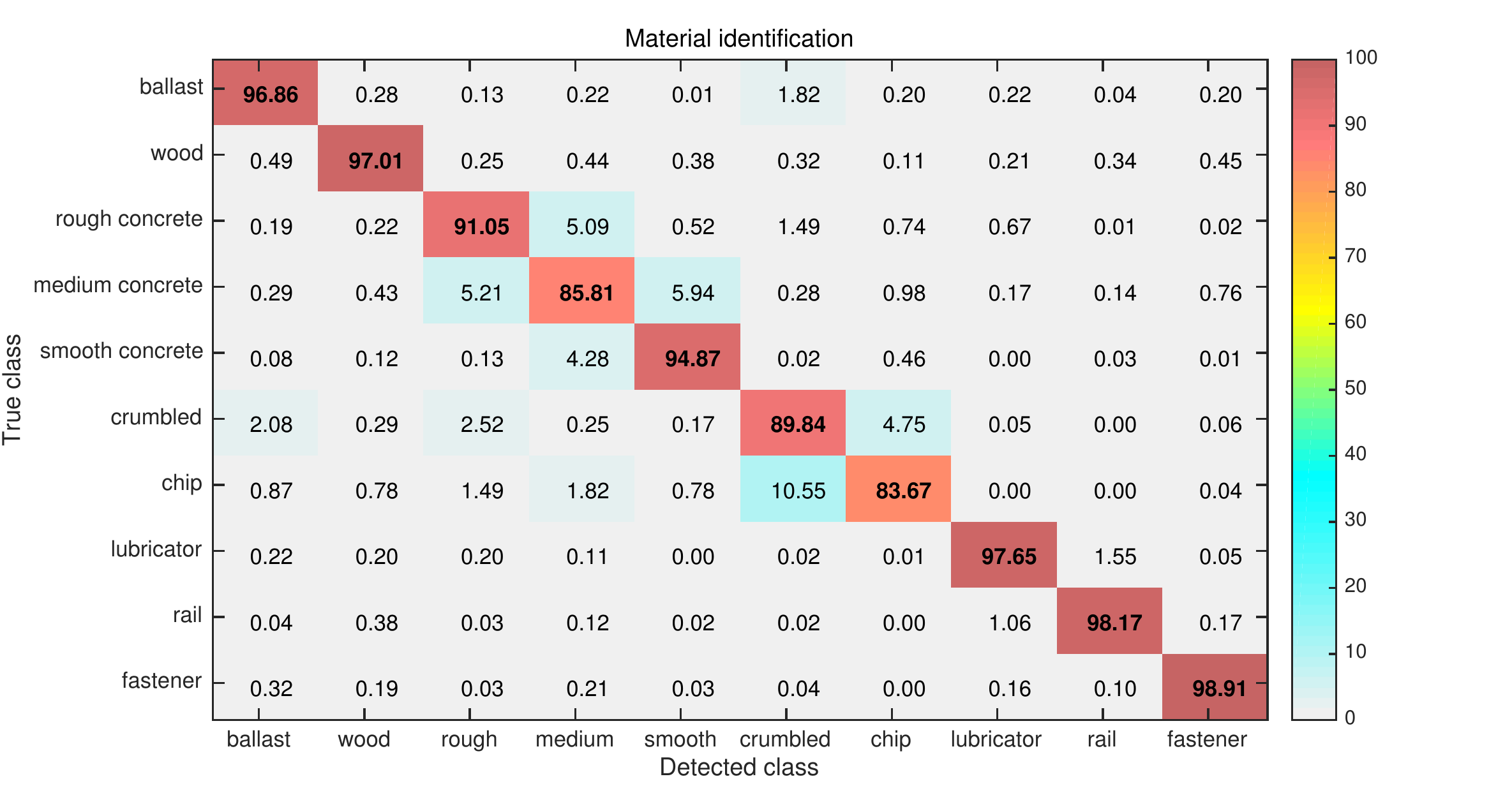} \\
  (a) & (b)
\end{tabular}
\end{center}
\caption
{ \label{fig:cm}
Confusion matrix of material classification on 2.5 million 80$\times$80 image patches with Deep Convolutional Neural Networks using (a) multi-task learning (b) single task learning\cite{GP15icip}.}
\end{figure*}

\subsection{Material Identification}

We divided the dataset into 5 splits and used 80\% of the images for training and 20\% for testing and we generated a model for each of the 5 possible training sets. For each split of the data, we randomly sampled 50,000 patches of each class. Therefore, for each model was trained with 2 million patches. We trained the network using a batch size of 128 for a total of 300,000 iterations with a momentum of 0.9 and a weight decay of $5 \times 10^{-5}$. The learning rate is initially set to 0.01 and it decays by a factor of 0.5 every 30,000 iterations.  The following methods are compared in this paper:

\begin{itemize}
\item \textbf{Deep CNN MTL 3:} The method described in Section~\ref{sec:material} with the full architecture in Figure~\ref{fig:architecture}.
\item \textbf{Deep CNN MTL 2:} The previous method without the binary SVM subnet.
\item \textbf{Deep CNN STL:} The previous method without the fasteners subnet and a batch size of 64. This single task learning baseline is exactly the same model used in \cite{GP15icip}.
\item \textbf{LBP-HF with approximate Nearest Neighbor:} The Local Binary Pattern Histogram Fourier descriptor introduced in \cite{ahonen2009rotation} is invariant to global image rotations while preserving local information. We used the implementation provided by the authors. To perform approximate nearest neighbor we used FLANN\cite{muja_flann_2009} with the 'autotune' parameter set to a target precision of 70\%.
\item \textbf{Uniform LBP with approximate Nearest Neighbor} The $LBP^{u2}_{8,1}$ descriptor \cite{ojala2002multiresolution} with FLANN.
\item \textbf{Gabor features with approximate Nearest Neighbor:} We filtered each image with a filter bank of 40 filters (4 scales and 8 orientations) designed using the code from \cite{haghighat2013identification}. As proposed in \cite{manjunath1996texture}, we compute the mean and standard deviation of the output of each filter and build a feature descriptor as $f = [ \mu_{00} \; \sigma_{00} \; y_{01} \; \dots \; \mu_{47} \; \sigma_{47}]$. Then, we perform approximate nearest neighbor using FLANN with the same parameters.
\end{itemize}

The material classification results are summarized in Table~\ref{table:classification_results} and the confusion matrices are shown in Figure~\ref{fig:cm}.
\begin{table}[htp!]
\caption{Material classification results.}
\small
\label{table:classification_results}
\begin{center}
\begin{tabular}{c | c }
  \hline
  Method & Accuracy \\
  \hline
  \hline
  Deep CNN MTL 3 & \textbf{95.02}\% \\
  Deep CNN MTL 2 & 93.60\% \\
  Deep CNN STL\cite{GP15icip} & 93.35\% \\
  LBP-HF with FLANN & 82.05\% \\
  $LBP^{u2}_{8,1}$ with FLANN & 82.70\% \\
  Gabor with FLANN & 75.63\% \\
  \hline
\end{tabular}
\end{center}
\end{table}

Since we are using a fully convolutional DCNN, we directly transfer the parameters learned using small patches to a network that takes one $4096 \times 320$ image as an input, and generates 10 score maps of dimension $252 \times 16$ each. The segmentation map is generated by taking the label corresponding to the maximum score. Figure~\ref{fig:example} shows several examples of concrete and wood ties, with and without defects and their corresponding segmentation maps.

\subsection{Crumbling and Chipped Tie Detection}

\begin{figure*}[htp!]
\begin{center}
\begin{tabular}{c}
  \includegraphics[trim=0mm 0mm 0mm 0mm, clip=true, width=3.5in]{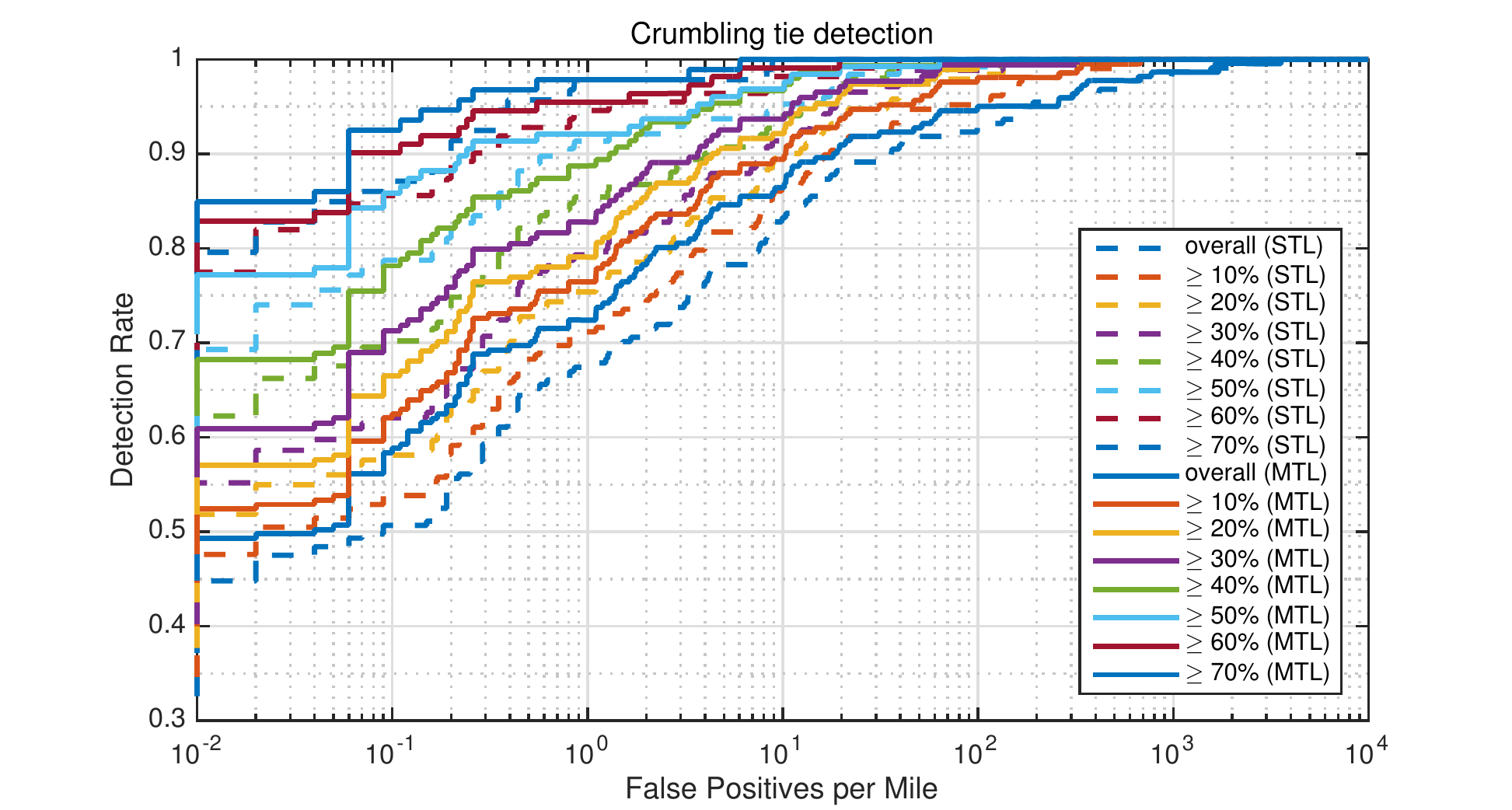} \includegraphics[trim=0mm 0mm 0mm 0mm, clip=true, width=3.5in]{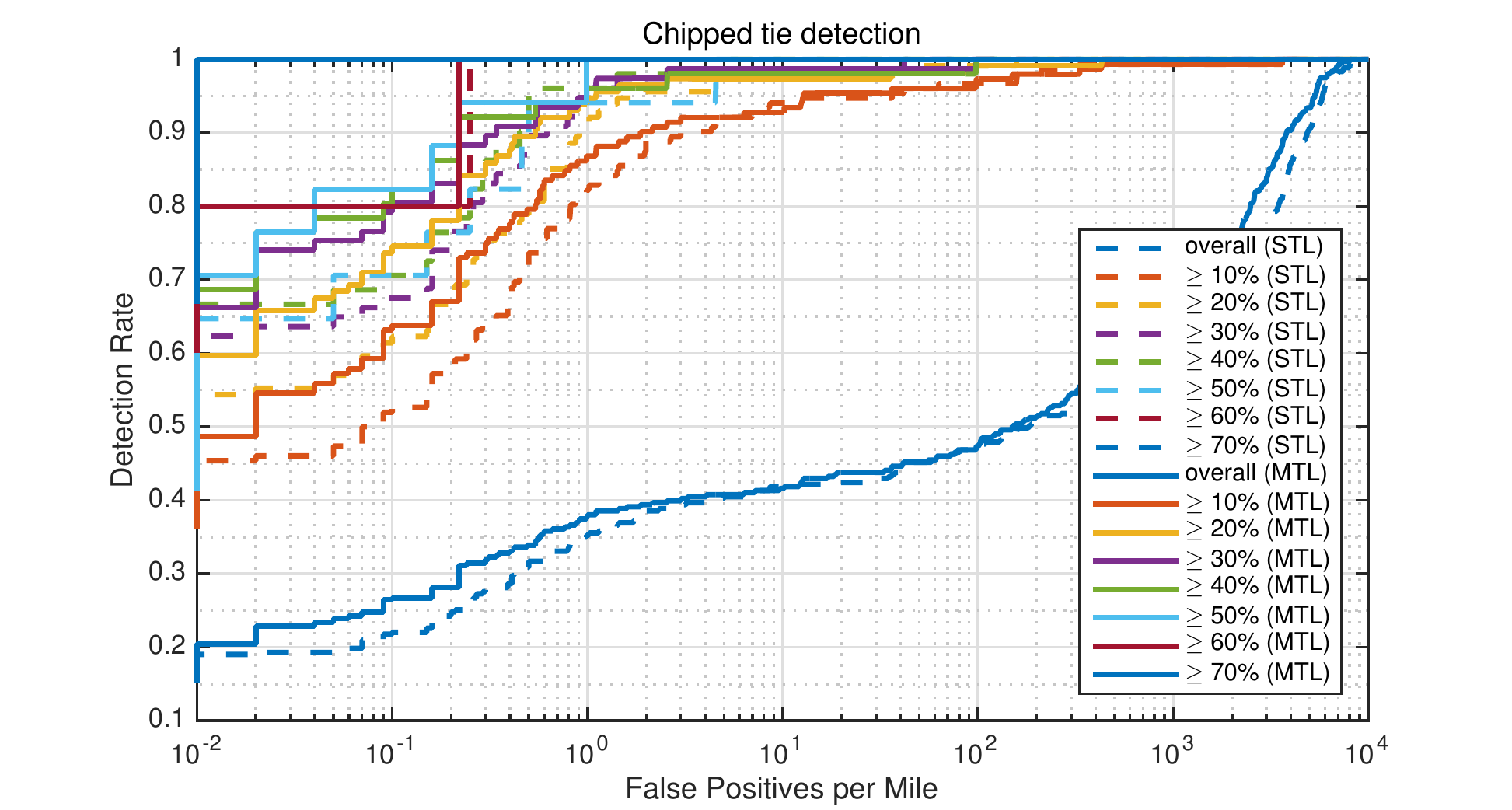} \\
  (a) \hskip 230pt (b) 
\end{tabular}
\end{center}\vskip 0pt
\caption
{ \label{fig:crumROC}
(a) ROC curve for detecting crumbling tie conditions. (a) ROC curve for detecting chip tie conditions. Each curve is generated considering conditions at or above a certain severity level. Note: False positive rates are estimated assuming an average of $10^4$ images per mile. Confusion between chipped and crumbling defects are not counted as false positives.}
\end{figure*}

\begin{figure}[htp!]
\begin{center}
\begin{tabular}{c c}
  \includegraphics[trim=0mm 0mm 0mm 0mm, clip=true, width=1.6in]{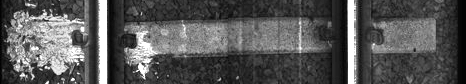} &
  \includegraphics[trim=0mm 0mm 0mm 0mm, clip=true, width=1.6in]{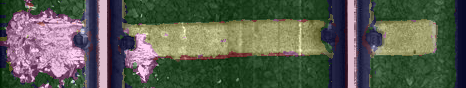} \\
  \includegraphics[trim=0mm 0mm 0mm 0mm, clip=true, width=1.6in]{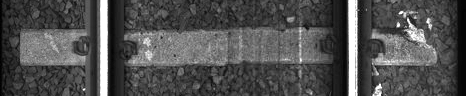} &
  \includegraphics[trim=0mm 0mm 0mm 0mm, clip=true, width=1.6in]{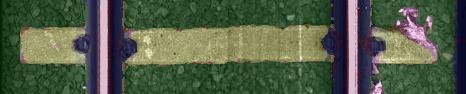} \\
  \includegraphics[trim=0mm 0mm 0mm 0mm, clip=true, width=1.6in]{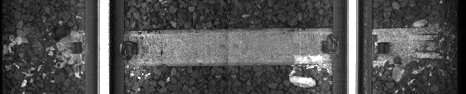} &
  \includegraphics[trim=0mm 0mm 0mm 0mm, clip=true, width=1.6in]{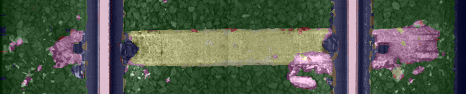} \\
  \includegraphics[trim=0mm 0mm 0mm 0mm, clip=true, width=1.6in]{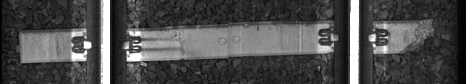} &
  \includegraphics[trim=0mm 0mm 0mm 0mm, clip=true, width=1.6in]{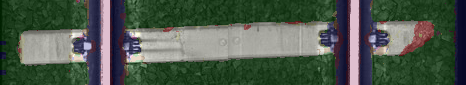} \\
  \includegraphics[trim=0mm 0mm 0mm 0mm, clip=true, width=1.6in]{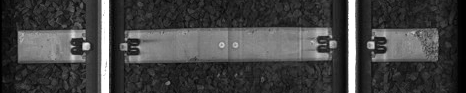} &
  \includegraphics[trim=0mm 0mm 0mm 0mm, clip=true, width=1.6in]{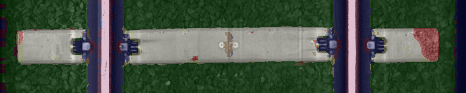} \\
  \includegraphics[trim=0mm 0mm 0mm 0mm, clip=true, width=1.6in]{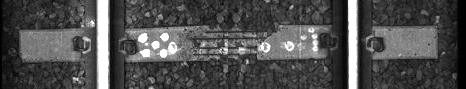} &
  \includegraphics[trim=0mm 0mm 0mm 0mm, clip=true, width=1.6in]{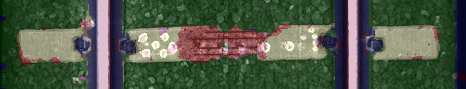} \\
  \includegraphics[trim=0mm 0mm 0mm 0mm, clip=true, width=1.6in]{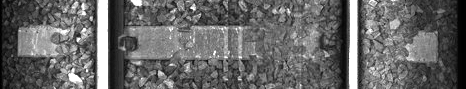} &
  \includegraphics[trim=0mm 0mm 0mm 0mm, clip=true, width=1.6in]{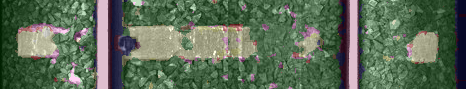} \\
  \includegraphics[trim=0mm 0mm 0mm 0mm, clip=true, width=1.6in]{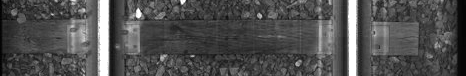} &
  \includegraphics[trim=0mm 0mm 0mm 0mm, clip=true, width=1.6in]{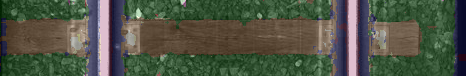} \\
\end{tabular}
\end{center}
\caption
{ \label{fig:example}
Semantic segmentation results (images displayed at 1/16 of original resolution). See Figure~\ref{fig:texture_classes} for color legend.
}
\end{figure}

The first 3 rows in Figure~\ref{fig:example} show examples of a crumbling ties and their corresponding segmentation map. Similarly, rows 4 through 6 show examples of chipped ties. To evaluate the accuracy of the crumbling and chipped tie detector described in Section \ref{sec:score} we divide each tie in 4 images and we evaluate the score \eqref{eq:scoret} on each image independently. Due to the large variation in the area affected by crumbling/chip we assigned a severity level to each ground truth defect, and for each severity level we plot the ROC curve of finding a defect when ignoring lower level defects. The severity levels are defined as the ratio of the inspectable area that is labeled as a defect. Figure~\ref{fig:crumROC} shows the ROC curves for each type of anomaly. Because of the choice of the fixed $\alpha=0.9$ in equation \eqref{eq:scoret} the performance is not reliable for defects under 10\% severity. For defects that are bigger than the 10\% threshold, at a false positive rate (FPR) of 10 FP/mile the true positive rates (TPR) are 89.42\% for crumbling and 93.42\% for chips. This is an improvement of 3.36\% and 1.31\% compared to the STL results reported in \cite{GP15icip}. The results on chipped tie detection are mixed, while there is an improvement at 2 FP/mile, the detection performance at 10 FP/mile is lower than that of STL. Table \ref{table:condition_results} summarizes the results.

\begin{table}[htp!]
\caption{Tie condition detection. For chipped and crumbling, only ties with at least 10\% affected area are included. Fastener rates correspond include those for which the track is clear.
}
\small
\label{table:condition_results}
\begin{center}
\begin{tabular}{c | c | c | c | c }
  \hline
  Condition & FPR & MTL & STL \\
  \hline
  \hline
  \multirow{ 2}{*}{Crumbling Tie} & 10 FP/mile & \textbf{89.42}\% & 86.54\% \\
  & 2 FP/mile & \textbf{82.21}\% & 74.52\% \\
  \hline
  \multirow{ 2}{*}{Chipped Tie} & 10 FP/mile & 92.76\% & \textbf{94.08}\% \\
   & 2 FP/mile & \textbf{90.13}\% & 88.52\% \\
  \hline
  \multirow{ 2}{*}{Fastener} & 10 FP/mile & \textbf{99.91}\% & 98.41\% \\
  & 2 FP/mile & \textbf{96.74}\% & 93.19\% \\
  \hline
\end{tabular}
\end{center}
\end{table}

\subsection{Fastener Categorization}

On our dataset, we have a total of 8 object categories (2 for broken clips, 1 for PR clips, 1 for e-clips, 2 for fast clips, 1 for c-clips, and 1 for j-clips) plus a special category for background (which includes missing fasteners). We also have 4 synthetically generated categories by mirroring non-symmetric object classes (PR, e, c, and j clips), so we use a total of 12 object categories at test time. 

\begin{figure*}[htp!]
\begin{center}
\setlength{\tabcolsep}{0pt}
\begin{tabular}{c c c}
  \includegraphics[trim=5.3mm 3mm 10mm 5mm, clip=true, width=2.3in]{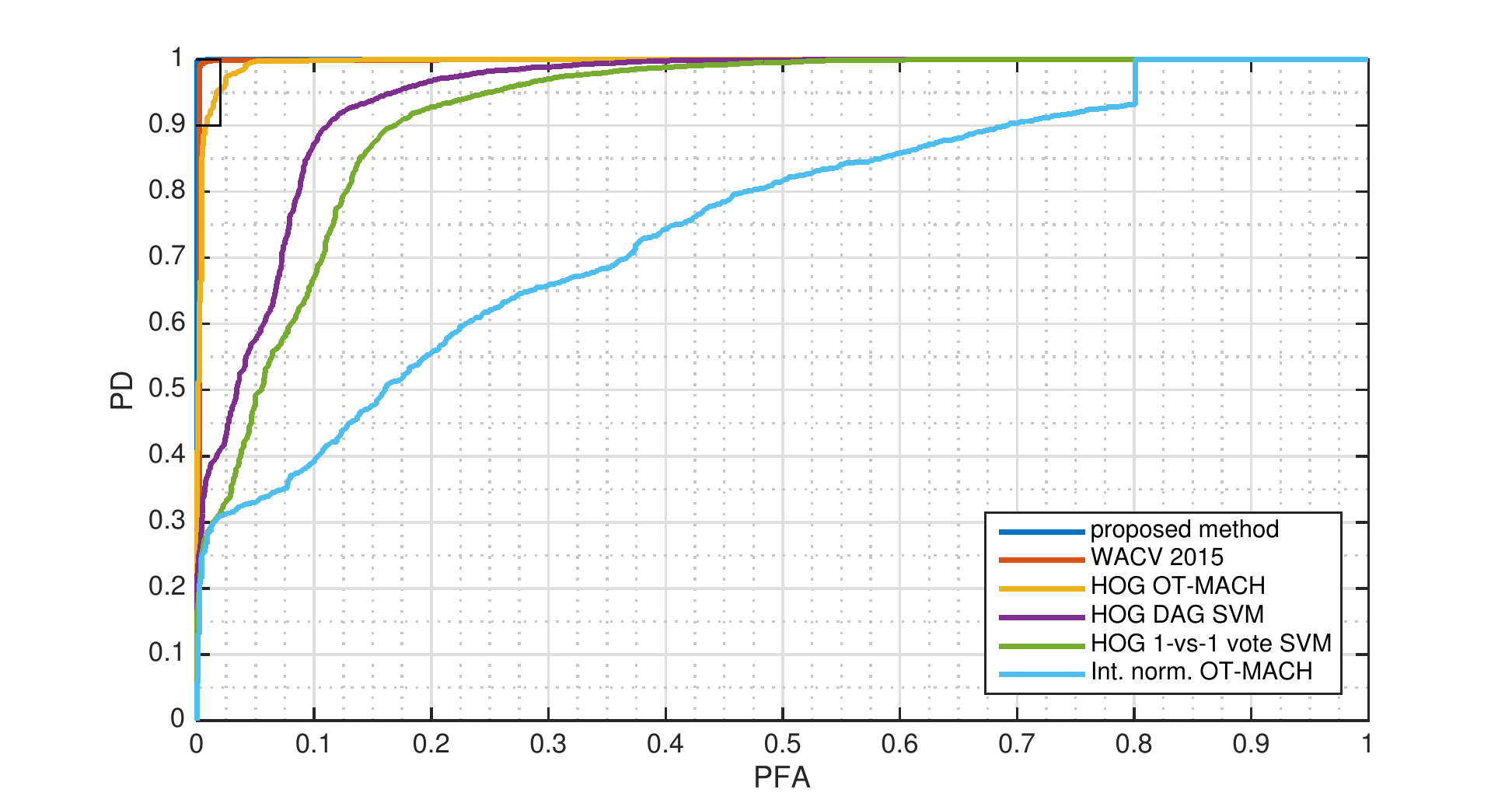} &
  \includegraphics[trim=5.3mm 3mm 10mm 5mm, clip=true, width=2.3in]{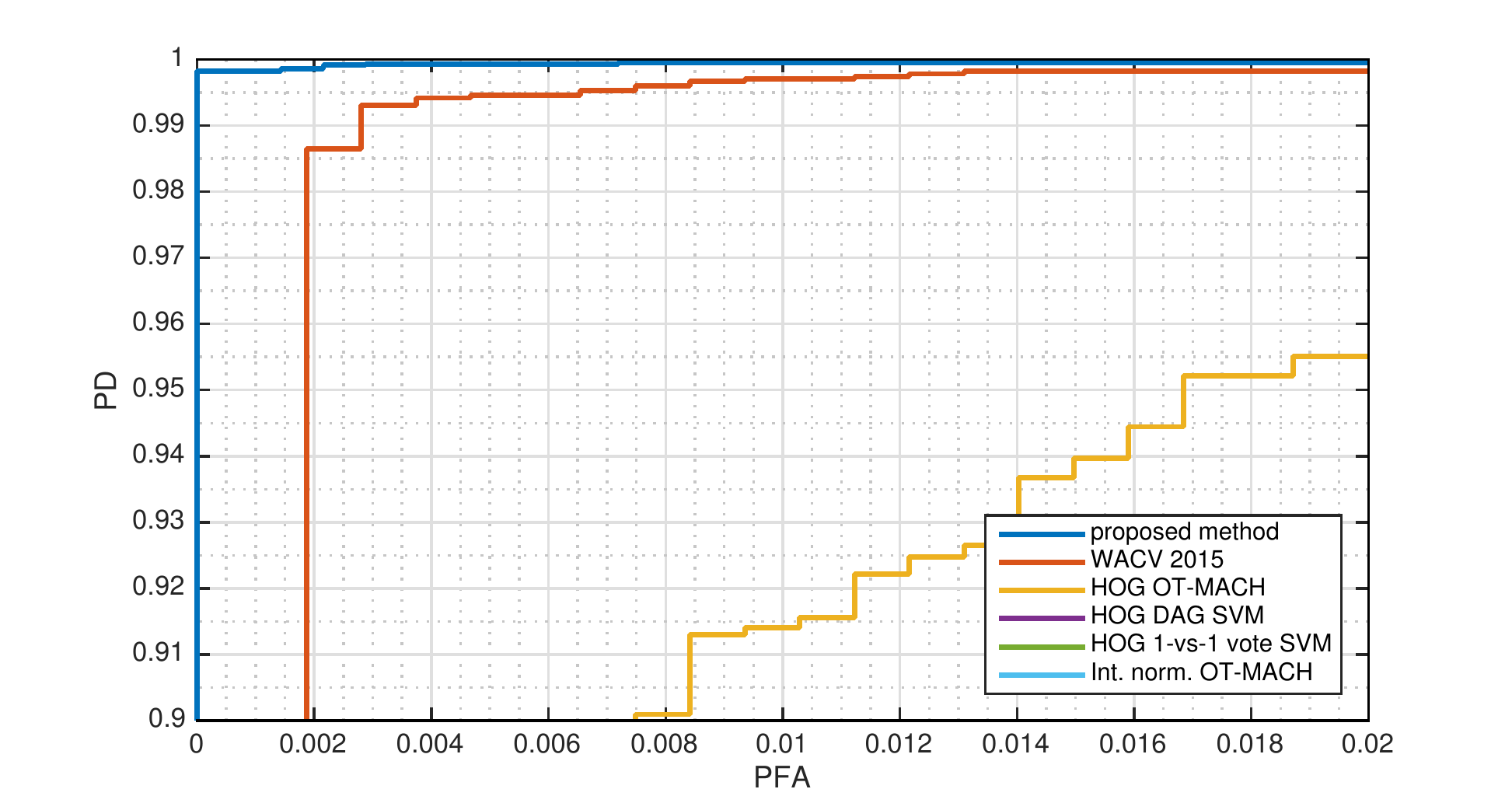} &
  \includegraphics[trim=5.3mm 3mm 10mm 5mm, clip=true, width=2.3in]{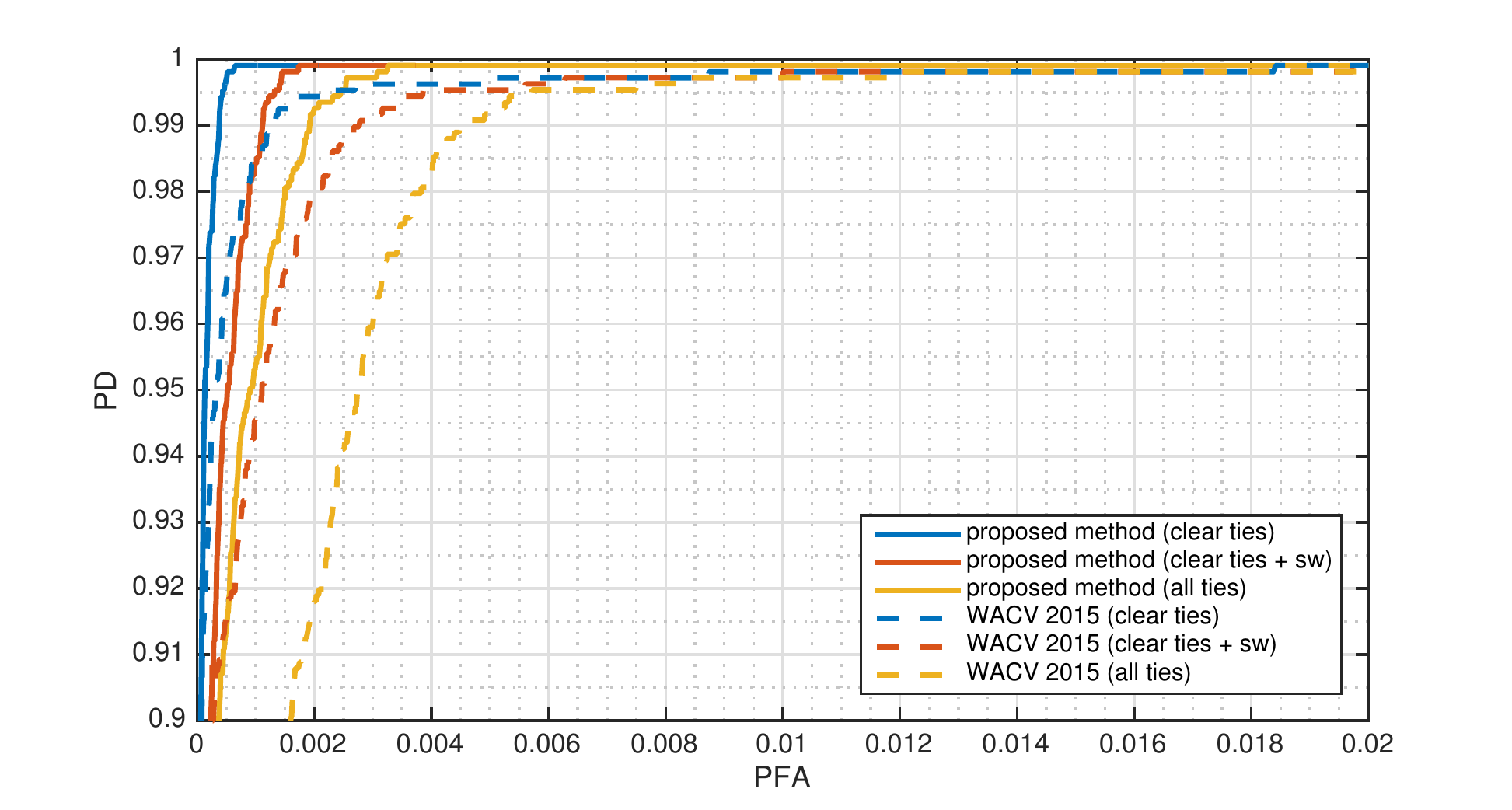} \\
  (a) & (a) detail & (b) \\
\end{tabular}
\end{center}
\caption
{ \label{fig:roc}
ROC curves for the task of detecting defective (missing or broken) fasteners (a) using 5-fold cross-validation on the training set (b) on the 85-mile testing set.
}
\end{figure*}

For training our detectors, we used  the same training set as in \cite{GP14fast}, which has a total of 3,805 image patches, including 1,069 good fasteners, 714 broken fasteners, 33 missing fasteners, and 1,989 patches of background texture. During training, we also included the mirrored versions of the missing/background patches and all symmetric object classes.

In addition to the proposed method described in Section \ref{sec:fastener}, we have also implemented and evaluated the following alternative methods:
\begin{itemize}
\item \textbf{STL (WACV 2015):} The method in \cite{GP14fast} uses the same scores as the proposed method, based on HOG features instead of the features learned at layer \emph{conv4\_f}.
\item \textbf{Intensity normalized OT-MACH:} As in \cite{PB09}, for each image patch, we subtract the mean and normalize the image vector to unit norm. For each class $c$, we design an OT-MACH filter in the Fourier domain using $h_c = [ \alpha I + (1-\alpha) D_c ]^{-1} \overline{x}_c$ with $\alpha = 0.95$, where $I$ is the identity matrix, $D_c = (1/N_c) \sum_{i=1}^{N_c} x_{ci} x_{ci}^*$, and $N_c$ is the number of training samples of class $c$.
\item \textbf{HOG features with OT-MACH:} The method in \cite{PB09}, but replacing intensity with HOG feature. Since HOG features are already intensity-invariant, the design of the filters reduces to $h_c=\overline{x}_c$.
\item \textbf{HOG features with DAG-SVM:} We run one-vs-one SVM classifiers in sequence. We first run each class against the background on each candidate region. If at least one classifier indicates that the patch is not background, then we run the DAG-SVM algorithm \cite{PCS00} over the remaining classes.
\item \textbf{HOG features with majority voting SVM:} We run all possible one-vs-one SVM classifiers and select the class with the maximum number of votes.
\end{itemize}
For the second and third methods, we calculate the score using the formulation introduced in sections \ref{sec:classification} and \ref{sec:scoref}, but with $b_c = h_c$ and $f_c = h_c - \sum_{i\neq c} h_i/(C-1)$. For the forth and last methods, we first estimate the most likely class in $\mathcal{G}$ and in $\mathcal{B}$. Then, we set $S_b$ as the output of the classifier between these two classes, and $S_m$ as the output of the classifier between the background and the most likely class.

We can observe in Figure \ref{fig:roc} (a) that the proposed method is the most accurate, followed by WACV 2015 STL baseline and HOG with OT-MACH method. The other methods perform poorly on this dataset.  In the third row of Table~\ref{table:condition_results} we compare the fastener detection performance of MTL with the STL baseline.


\subsection{Defect Detection}

To evaluate the performance of our defect detector, we divided each tie into 4 regions of interest (left field, left gage, right gage, right field) and calculated the score defined by (\refeq{eq:scoref}) for each of them. Figure \ref{fig:roc} shows the ROC curve for crossvalidation on the training set as well as for the testing set of 813,148 ROIs (203,287 ties). The testing set contains 1,052 ties images with at least one defective fastener per tie. The total number of defective fasteners in the testing set was 1,087 (0.13\% of all the fasteners), including 22 completely missing fasteners and 1,065 broken fasteners. The number of ties that we flagged as ``uninspectable'' is 2,524 (1,093 on switches, 350 on lubricators, 795 covered in ballast, and 286 with other issues).

We used the ROC on clear ties (blue curve) in Figure \ref{fig:roc} (b) to determine the optimal threshold to achieve a design false alarm rate of $0.07\%$ ($\tau = 0.1070$). This target is a bit lower than the $0.1\%$ that we used in the for the baseline experiments. The reason for lowering the sensitivity is that the detection rate pateaus at $\text{PFA} >0.06\%$. Using this sensitivity level, we ran our defective fastener detector at the tie level (by taking the minimum score across all 4 regions).  Results are shown in Table \ref{table:tie_results}.

\begin{table}[htp!]
\caption{Results for detection of ties with at least one defective fastener.}
\small
\label{table:tie_results}
\begin{center}
\setlength{\tabcolsep}{0.4em}
\begin{tabular}{c | c | c | c | c | c | c}
  \hline
  \multirow{2}{*}{Subset} & \multirow{2}{*}{Total} & \multirow{2}{*}{\# Bad} & \multicolumn{2}{c|}{PD} & \multicolumn{2}{c}{PFA} \\
  & & & MTL & STL & MTL & STL \\
  \hline
  \hline
  clear ties & 200,763 & 1,037 & \textbf{99.90\%} & 98.36\% & \textbf{0.25\%} & 0.38\% \\
  clear + sw. & 201,856 & 1,045 & \textbf{99.90\%} & 97.99\% & \textbf{0.61\%} & 0.71\% \\
  all ties & 203,287 & 1,052 & \textbf{99.90\%} & 98.00\% & \textbf{1.01\%} & 1.23\% \\
  \hline
\end{tabular}
\end{center}
\end{table}

\begin{figure}[htp!]
\begin{center}
  \includegraphics[width=3.25in]{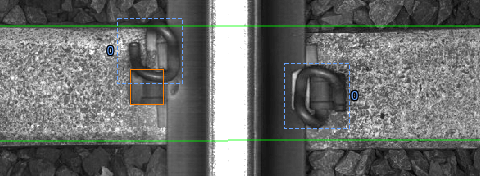}
\end{center}
\caption
{ \label{fig:single_missed}
The single defect missed by our detector. Solid bounding boxes correspond to ground truth annotations. Dashed bounding boxes correspond to the output of the detector. The number 0 corresponds to the PR-clip class, which is correctly classified. The clip has not completely popped out.}
\end{figure}

At this sensitivity level, our MTL detector only misses one defect (compared to 17 type II errors with the baseline detector). The false alarm rate on clear ties goes down to $0.25\%$, which is $34\%$ lower than the baseline. Figure \ref{fig:single_missed} shows the single defective fastener that was missed. It could be argued that the clip is still holding the rail in place, so it is a very close call.

\section{Conclusion and Future Work} \label{sec:conclusion}

This paper has introduced a new algorithm for inspecting railway ties and fasteners that takes advantage of the inherent structure of this problem. We have been able to benefit from scalability advantage of deep convolutional neural networks despite the limited amount of training data in some of the classes. This has been possible by setting up multiple tasks and cooperatively training a shared representation that is effective on each of them. We have showed that not only is possible save computation time by reusing the computation of intermediate features, but also that this representation results in better generalization performance than traditional features.


%



\section*{Acknowledgment}

This work was partially supported by the Federal Railroad Administration under contract DTFR53-13-C-00032. The authors thank Amtrak, ENSCO, Inc. and the Federal Railroad Administration for providing the data used in this paper. The authors sincerely thank University of Maryland student Daniel Bogachek for his help setting up earlier crumbling tie detection experiments during his visit at the Center for Automation Research in summer 2014.

\ifCLASSOPTIONcaptionsoff
  \newpage
\fi



\bibliographystyle{IEEEtran}
\bibliography{ShearGPU_ASP_2013,deeptrackbib}
\end{document}